\documentclass[journal]{IEEEtran} 
\usepackage{etex}
\usepackage{amsthm,array,  multirow, bm, amsmath, mathtools, amssymb,booktabs, subcaption}  
\usepackage[bookmarks=false, colorlinks=true, allcolors=blue]{hyperref}  
\usepackage{algorithm}

\usepackage{algorithm, algorithmic}
\usepackage{cite}
\usepackage{chngcntr}

\newcommand{\movobjgrouseFrames}{\includegraphics[scale=0.995, trim={5.5cm, 0.4cm, 5.35cm, 1.1cm}, clip]{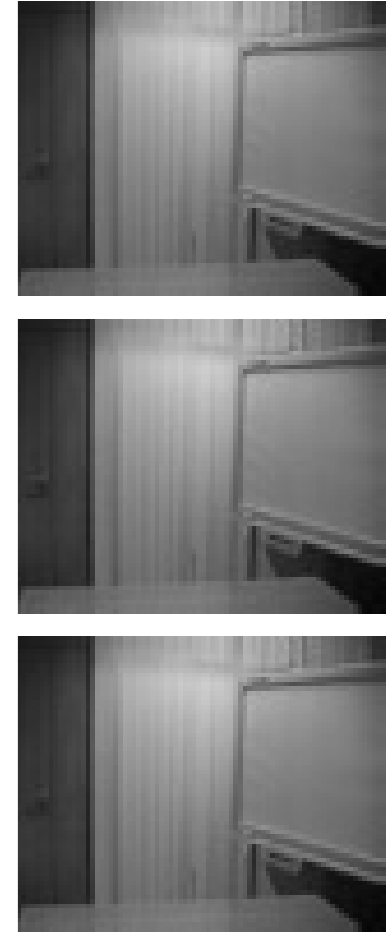}}

\newcommand{\movobjorigFrames}{\includegraphics[scale=0.995, trim={5.5cm, 0.4cm, 5.35cm, 1.1cm}, clip]{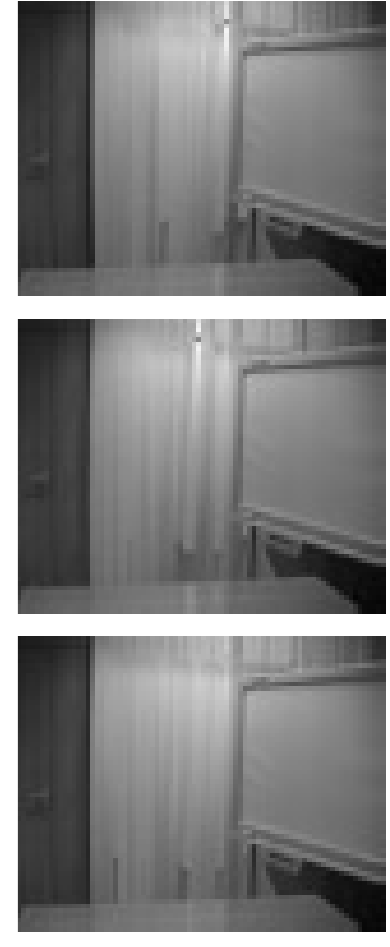}}
\newcommand{\movobjcorruptedFrames}{\includegraphics[scale=1.015, trim={5.69cm, 0.4cm, 5.5cm, 1.1cm}, clip]{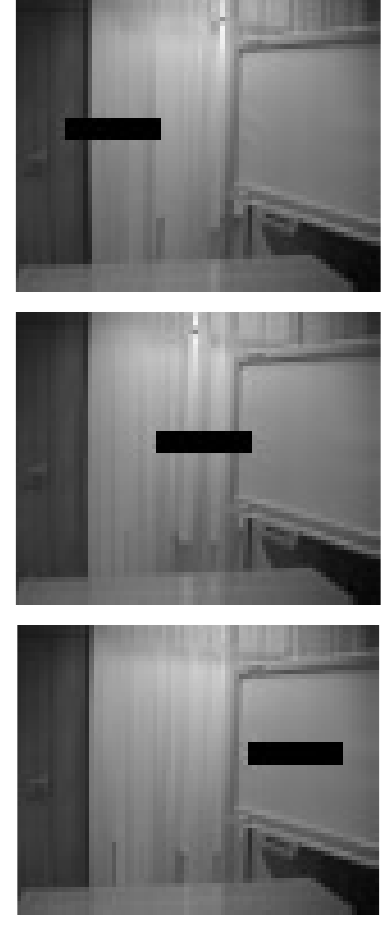}}
\newcommand{\movobjnorstFrames}{\includegraphics[scale=1, trim={5.5cm, 0.4cm, 5.5cm, 1.1cm}, clip]{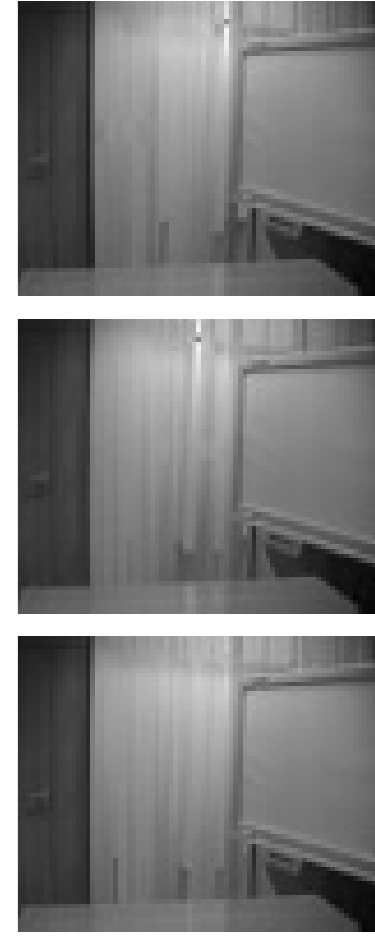}}
\newcommand{\movobjpetrelsXFrames}{\includegraphics[scale=1, trim={5.50cm, 0.4cm, 5.6cm, 1.cm}, clip]{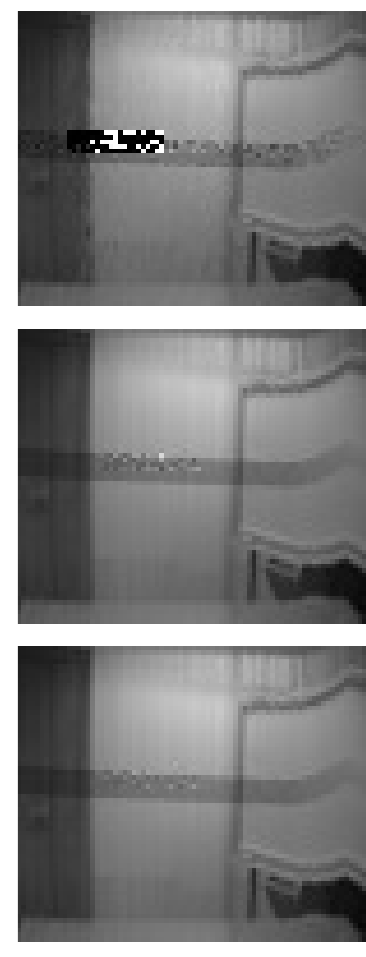}}
\newcommand{\movobjialmFrames}{\includegraphics[scale=1, trim={5.5cm, 0.4cm, 5.5cm, 1.1cm}, clip]{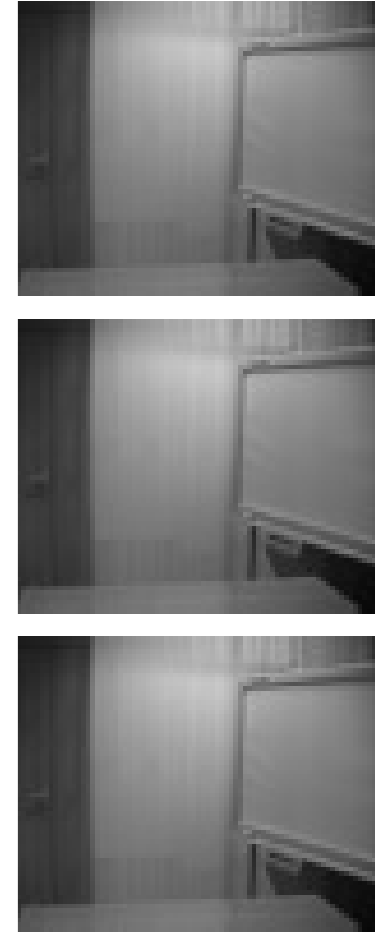}}
\newcommand{\movobjsvtFrames}{\includegraphics[scale=1, trim={5.5cm, 0.4cm, 5.5cm, 1.1cm}, clip]{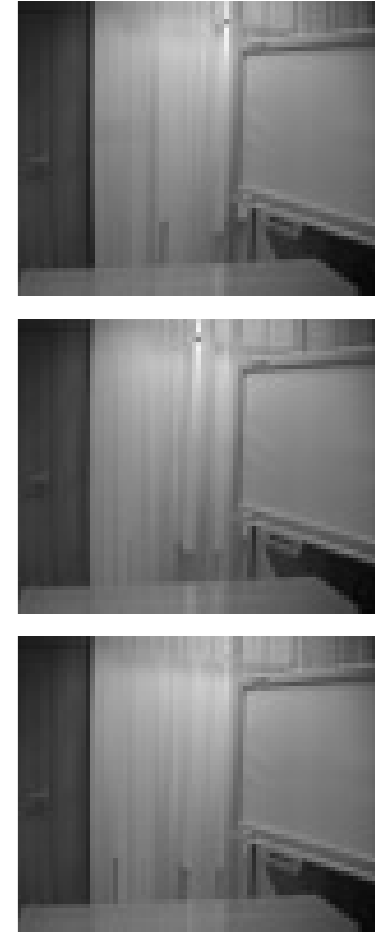}}

\newcommand{\bgfgorigFrames}{\includegraphics[scale=0.995, trim={5.5cm, 0.4cm, 5.35cm, 1.1cm}, clip]{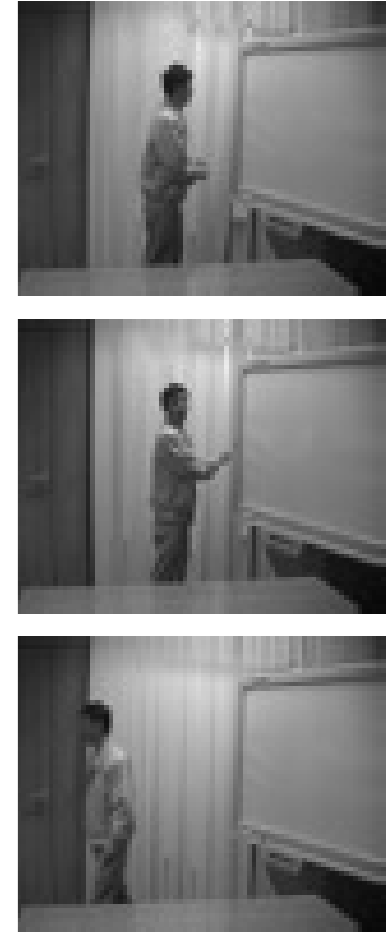}}
\newcommand{\bgfgcorruptedFrames}{\includegraphics[scale=0.995, trim={5.5cm, 0.4cm, 5.35cm, 1.1cm}, clip]{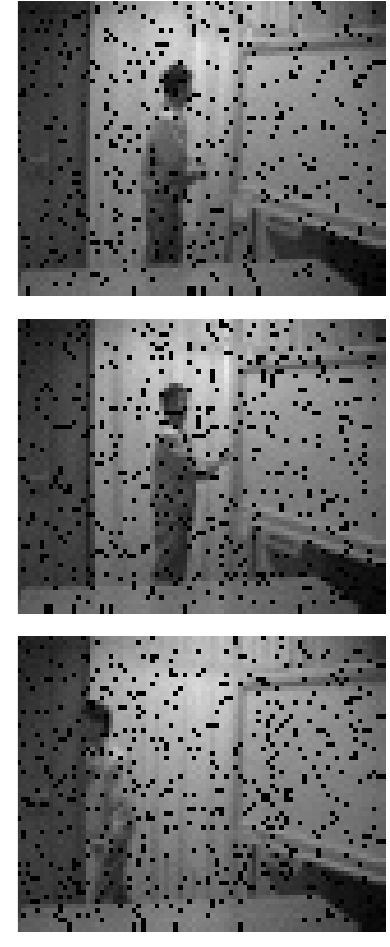}}
\newcommand{\bgfgNORSTFrames}{\includegraphics[scale=1.01, trim={5.65cm, 0.4cm, 5.5cm, 1.1cm}, clip]{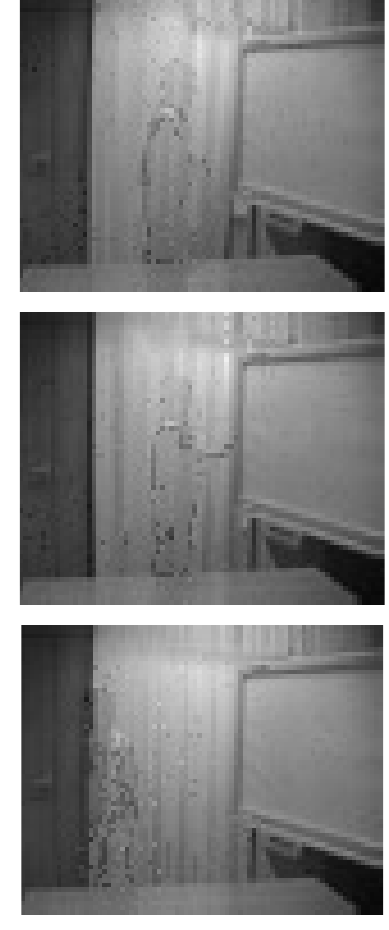}}
\newcommand{\bgfgGRASTAFrames}{\includegraphics[scale=0.995, trim={5.5cm, 0.4cm, 5.35cm, 1.1cm}, clip]{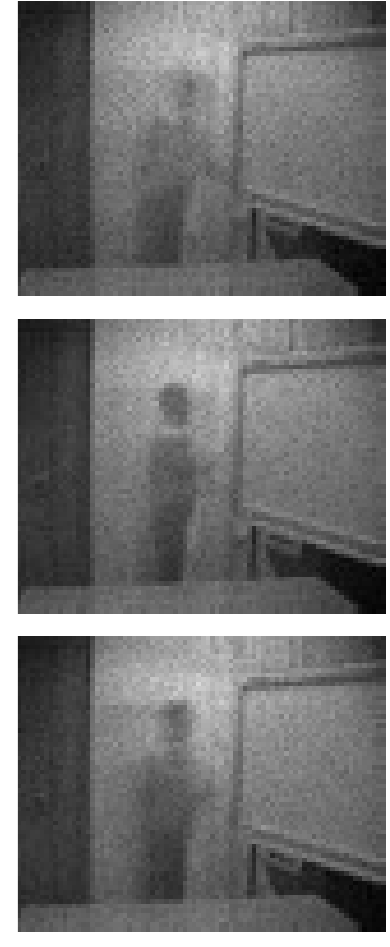}}
\newcommand{\bgfgNCRMCFrames}{\includegraphics[scale=0.995, trim={5.5cm, 0.4cm, 5.35cm, 1.1cm}, clip]{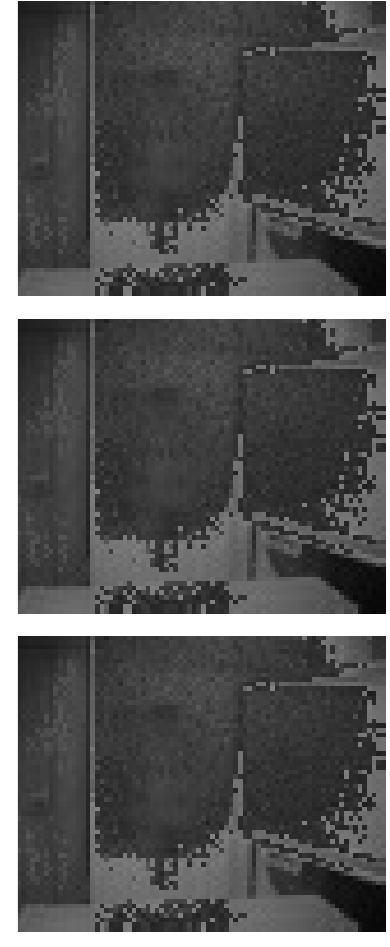}}

\renewcommand{\L}{\bm{L}}

\newcommand{\norm}[1]{\left\|#1\right\|}

\newtheorem{theorem}{Theorem}

\newtheorem{corollary}[theorem]{Corollary}
\newtheorem{definition}[theorem]{Definition}
\newtheorem{remark}[theorem]{Remark}

\newcommand{\bi}{\begin{itemize}}
\newcommand{\ei}{\end{itemize}}
\newcommand{\ben}{\begin{enumerate}}
\newcommand{\een}{\end{enumerate}}

\newcommand{\bean}{\begin{eqnarray*} }
\newcommand{\eean}{\end{eqnarray*} }
\newcommand{\bea}{\begin{eqnarray} }
\newcommand{\eea}{\end{eqnarray} }
\newcommand{\ba}{\begin{align*} }
\newcommand{\ea}{\end{align*} }

\newcommand{\outfracrow}{\small{\text{max-outlier-frac-row}}}
\newcommand{\outfraccol}{\small{\text{max-outlier-frac-col}}}

\newcommand{\dif}{{\text{dif}}}
\newcommand{\xhat}{\bm{\hat{x}}}

\newcommand{\bl}{\begin{frame}}
\newcommand{\el} {\end{frame}}

\renewcommand\thetheorem{\arabic{section}.\arabic{theorem}}

\newcommand{\tmax}{d} 

\newcommand{\wt}{\bm{w}_t}

\newcommand{\xt}{\bm{x}_t}
\newcommand{\x}{\bm{x}}

\renewcommand{\l}{\bm{\ell}}

\newcommand{\lt}{\bm{\ell}_t}
\newcommand{\lhat}{\hat{\bm{\ell}}}

\newcommand{\lhatt}{\hat{\l}_t}   
\newcommand{\yt}{\bm{y}_t}
\newcommand{\y}{\bm{y}}
\newcommand{\tty}{\tilde{\bm{y}}}
\newcommand{\w}{\bm{w}}
\renewcommand{\v}{{\bm{\nu}}}
\newcommand{\vt}{\v_t}

\renewcommand{\a}{\bm{a}}

\newcommand{\et}{\bm{e}_t}

\newcommand{\at}{\bm{a}_t}

\newcommand{\I}{\bm{I}}

\newcommand{\Lam}{\bm{\Lambda}}

\newcommand{\T}{\mathcal{T}}
\newcommand{\J}{\mathcal{J}}

\newcommand{\Lhat}{\hat{\bm{L}}}

\renewcommand{\P}{\bm{P}}
\newcommand{\U}{\bm{U}}
\newcommand{\V}{\bm{V}}

\newcommand{\Phat}{\hat{\bm{P}}}

\newcommand{\Span}{\operatorname{span}} 

\newcommand{\basis}{\operatorname{basis}}
\newcommand{\rank}{\operatorname{rank}}
\newcommand{\E}{\mathbb{E}}

\newcommand{\train}{\mathrm{train}}
\newcommand{\That}{\hat{\mathcal{T}}}

\newcommand{\SE}{\sin\theta_{\max}}

\newcommand{\that}{{\hat{t}}}

\newcommand{\M}{\bm{M}}

\newcommand{\X}{\bm{X}}
\newcommand{\Y}{\bm{Y}}

\newcommand{\one}{\bm{1}}

\newcommand{\B}{\bm{B}}
\renewcommand{\Re}{\mathbb{R}}

\newcolumntype{C}[1]{>{\centering\let\newline\\\arraybackslash\hspace{0pt}}m{#1}}

\newcommand{\bz}{b} 
\newcommand{\rmat}{r_{\mat}}

\newcommand{\z}{\bm{z}}

\newcommand{\ep}{\mathbb{E}}

\newcommand{\bt}{\bm{b}_t}
\newcommand{\zt}{\bm{Z}_t}

\newcommand{\zz} {\varepsilon} 

\newcommand{\Tt}{\mathcal{T}_t}

\newcommand{\smin}{x_{\min}}

\newcommand{\pt}{\bm{P}}

\newcommand{\bphi}{\bm{\Phi}}
\newcommand{\bpsi}{\bm{\Psi}}

\newcommand{\lthres}{\omega_{evals}} 

\newcommand{\tildej}{j}

\renewcommand{\zz}{\epsilon}
\usepackage{tikz}
\usepackage{pgfplots}
\usepgfplotslibrary{groupplots}
\usetikzlibrary{spy}

\pgfplotsset{select coords between index/.style 2 args={
    x filter/.code={
        \ifnum\coordindex<#1\fi
        \ifnum\coordindex>#2\fi
    }
}}

\pgfplotstableread[col sep = comma]{plots/final_files/bern_rho10.dat}\bernfix
\pgfplotstableread[col sep = comma]{plots/final_files/change_lambda_log.dat}\changelambda
\pgfplotstableread[col sep = comma]{plots/final_files/mo_rho20_new.dat}\mofix

\pgfplotstableread[col sep = comma]{plots/final_files/changing_ss_noise.dat}\changenoise
\pgfplotstableread[col sep = comma]{plots/final_files/changing_ss_eachtime.dat}\changetime
\pgfplotstableread[col sep = comma]{plots/final_files/pwconst_just.dat}\pwjust
\pgfplotstableread[col sep = comma]{plots/final_files/pwconst_just_all.dat}\pwjustall
\renewcommand{\zt}{\bm{z}_t}

\newcommand{\xmint}{x_{\min}}

\renewcommand{\dif}{\Delta}
\renewcommand{\rmat}{r_{\scriptscriptstyle{L}}}
\renewcommand{\S}{\bm{X}}
\renewcommand{\SE}{\mathrm{SE}}

\newcommand{\g}{\bm{g}}
\newcommand{\gt}{\g_t}

\renewcommand{\outfracrow}{\small{\text{max-out-frac-row}}}
\renewcommand{\outfraccol}{\small{\text{max-out-frac-col}}}

\newcommand{\sparse}{\mathrm{sparse}}

\newcommand{\proj}{\mathcal{P}_{\Omega_t}}

\newcommand{\Tmisst}{{\T_{t}}}
\newcommand{\Tspart}{{\T_{\sparse,t}}}
\newcommand{\missfracrow}{\small{\text{max-miss-frac-row}}}
\newcommand{\missfraccol}{\small{\text{max-miss-frac-col}}}
\newcommand{\tP}{\tilde{\P}}

\newcommand{\fracobs}{\rho}

\renewcommand{\one}{\mathbf{1}}

\renewcommand{\SE}{\mathrm{dist}}

\newcommand{\Subsection}[1]{ \vspace{-0.05in} \subsection{#1}  \vspace{-0.05in} }
\begin{document}

\title{Subspace Tracking from Errors and Erasures}
\title{Provable Subspace Tracking from Missing Data and Matrix Completion} 
\author{Praneeth Narayanamurthy,~\IEEEmembership{Student Member,~IEEE,} Vahid Daneshpajooh, and Namrata Vaswani,~\IEEEmembership{Fellow,~IEEE}
\thanks{Parts of this paper will be presented at IEEE International Conference on Acoustics, Speech, and Signal Processing, 2019 \cite{rst_miss_icassp} and IEEE International Symposium on Information Theory, 2019 \cite{st_miss_isit}.}
\thanks{The authors are with Department of Electrical and Computer Engineering, Iowa State University, Ames,
IA, 50010 USA (e-mail: {\{pkurpadn, vahidd, namrata\} @iastate.edu}).}
}
\maketitle

\begin{abstract}
We study the problem of subspace tracking in the presence of missing data (ST-miss). In recent work, we studied a related problem called robust ST. In this work, we show that a simple modification of our robust ST solution also provably solves ST-miss and robust ST-miss. To our knowledge, our result is the first ``complete'' guarantee for ST-miss. This means that we can prove that under assumptions on only the algorithm inputs, the output subspace estimates are close to the true data subspaces at all times. Our guarantees hold under mild and easily interpretable assumptions, and allow the underlying subspace to change with time in a piecewise constant fashion.  In contrast, all existing guarantees for ST are partial results and assume a fixed unknown subspace. Extensive numerical experiments are shown to back up our theoretical claims. Finally, our solution can be interpreted as a provably correct mini-batch and memory-efficient solution to low rank Matrix Completion (MC). 
\end{abstract}


\renewcommand{\subsubsection}[1]{{\bf #1. }}


\section{Introduction}
Subspace tracking from missing data (ST-miss) is the problem of tracking the (fixed or time-varying) low-dimensional subspace in which a given data sequence approximately lies when some of the data entries are not observed. The assumption here is that consecutive subsets of the data are well-approximated as lying in a subspace that is significantly lower-dimensional than the ambient dimension. Time-varying subspaces is a more  appropriate model for long data sequences (e.g. long surveillance videos). For such data, if a fixed subspace model is used, the required subspace dimension may be too large. As is common in time-series analysis, the simplest model for time-varying quantities is to assume that they are piecewise constant with time. We adopt this model here. If the goal is to provably track the subspaces to any desired accuracy, $\zz>0$, then, as we explain later in Sec. \ref{identif},  this assumption is, in fact, necessary.
{
Of course, experimentally, our proposed algorithm, and all existing ones, ``work'' (return good but not perfect estimates) even without this assumption, as long as the amount of change at each time is small enough. The reason is one can interpret subspace changes at each time as a ``piecewise constant subspace'' plus noise. The algorithms are actually tracking the ``piecewise constant subspace'' up to the noise level. We explain this point further in Sec. \ref{identif}.
}



ST-miss can be interpreted as an easier special case of robust ST (ST in the presence of additive sparse outliers) \cite{rrpcp_icml}. {We also study robust ST-miss which is a generalization of both ST-miss and robust ST.}  
Finally, our solutions for ST-miss and robust ST-miss also provide novel mini-batch solutions for low-rank matrix completion (MC) and robust MC respectively. 


Example applications where these problems occur include recommendation system design and video analytics. %
In video analytics, foreground occlusions are often the source of both missing and corrupted data: if the occlusion is easy to detect by simple means, e.g., color-based thresholding, then the occluding pixel can be labeled as ``missing''; while if this cannot be detected easily, it is labeled as an outlier pixel. Missing data also occurs due to detectable video transmission errors (typically called ``erasures'').
In recommendation systems, data is missing because all users do not label all items. 
In this setting, time-varying subspaces model the fact that, as different types of users enter the system, the factors governing user preferences change. 

\subsubsection{Brief review of related work}
ST has been extensively studied in both the controls' and the signal processing literature, see \cite{golubtracking, chi_review, sslearn_jmlr, rrpcp_proc} for comprehensive overviews of both classical and modern approaches. Best known existing algorithms for ST and ST-miss include Projection Approximate Subspace Tracking (PAST) \cite{past,past_conv}, Parallel Estimation and Tracking by Recursive Least Squares (PETRELS) \cite{petrels} and Grassmannian Rank-One Update Subspace Estimation (GROUSE) \cite{grouse,local_conv_grouse, grouse_global, grouse_enh}.  Of these, PETRELS is known to have the best experimental performance. There have been some attempts to obtain guarantees for GROUSE and PETRELS for ST-miss \cite{local_conv_grouse,grouse_global, petrels_new}, however all of these results assume the statistically stationary setting of a {\em fixed unknown subspace} and all of them provide only {\em partial guarantees.} This means that the result does not tell us what assumptions the algorithm inputs  (input data and/or initialization) need to satisfy in order to ensure that the algorithm output(s) are close to the true value(s) of the quantity of interest, either at all times or at least at certain times.
For example, \cite{local_conv_grouse} requires that the intermediate algorithm estimates of GROUSE need to satisfy certain properties (see Theorem \ref{thm:grouse} given later). It does not tell us what assumptions on algorithm inputs will ensure that these properties hold. On the other hand, \cite{petrels_new} guarantees closeness of the PETRELS output to a quantity other than the true value of the ``quantity of interest'' (here, the true data subspace); see Theorem \ref{thm:petrels}. Of course,  the advantage of GROUSE and PETRELS is that they are streaming solutions (require a single-pass through the data). This may also be the reason that a complete guarantee is harder to obtain for these.
Other related work includes streaming PCA with missing data \cite{streamingpca_miss,eldar_jmlr_ss}. A provable algorithmic framework for robust ST is Recursive Projected Compressive Sensing (ReProCS) \cite{rrpcp_allerton,rrpcp_perf,rrpcp_aistats,rrpcp_dynrpca,rrpcp_icml}. 
Robust ST-miss has not received much attention in the literature.

Provable MC has been extensively studied, e.g., \cite{matcomp_candes,lowrank_altmin,rmc_gd}. We discuss these works in detail in Sec. \ref{sec:prior_art}.


\subsubsection{Contributions}
(1) We show that a simple modification of a ReProCS-based algorithm called Nearly Optimal Robust ST via ReProCS (NORST for short) \cite{rrpcp_icml} also provably solves the ST-miss problem while being fast and memory-efficient. An extension for robust ST-miss is also presented.
Unlike all previous work on ST-miss, our guarantee is a {\em complete guarantee (correctness result)}: we show that, with high probability (whp), under simple assumptions on only the algorithm inputs, the output subspace estimates are close to the true data subspaces and get to within $\zz$ accuracy of the current subspace within a ``near-optimal'' delay. Moreover, unlike past work, our result allows time-varying subspaces (modeled as piecewise-constant with time) and shows that NORST-miss can provably detect and track each changed subspace quickly. Here and below, {\em near-optimal} means that our bound is within logarithmic factors of the minimum required. For $r$-dimensional subspace tracking, the minimum required delay is $r$; thus our delay of order $r\log n \log(1/\zz)$ is {\em near-optimal}.
Moreover, since ST-miss is an easier problem than robust ST, our guarantee for ST-miss is significantly better than the original one  \cite{rrpcp_icml} that it follows from. It does not assume a good first subspace initialization and does not require slow subspace change.

(2) Our algorithm and result can also be interpreted as a novel provably correct mini-batch and memory-efficient solution to low rank MC. We explain in Sec. \ref{sec:mainres_noisefree} that our guarantee is particularly interesting in the regime when subspace changes frequently enough, e.g., if it changes every order $r\log n \log(1/\zz)$ time instants.

\subsubsection{Organization}
We explain the algorithm and provide the guarantees for it in Sec. \ref{sec:norstmiss}; first for the noise-free case and then for the noisy case. A detailed discussion is also given that explains why our result is an interesting solution for MC.
In this section, we also develop simple heuristics that improve the experimental performance of NORST-miss.
We provide a detailed discussion of existing guarantees and how our work relates to the existing body of work in Sec. \ref{sec:prior_art}. Robust ST-miss is discussed in Sec. \ref{sec:norstmissrob}. Exhaustive experimental comparisons for simulated and partly real data (videos with simulated missing entries) are provided in Sec.  \ref{sec:sims_main}. These show that as long as the fraction of missing entries is not too large, (i) basic NORST-miss is nearly as good as the best existing ST-miss approach (PETRELS), while being faster and having a {\em complete guarantee}; (ii) its extensions have better performance than PETRELS and are also faster than PETRELS;  (iii) the performance of NORST-miss is worse than convex MC solutions, but much better than non-convex ones (for which code is available); however, NORST-miss is much faster than the convex MC methods.
We conclude in Sec. \ref{sec:conc}. 

\Subsection{Notation}
We use the interval notation $[a, b]$ to refer to all integers between $a$ and $b$, inclusive, and we use $[a,b): = [a,b-1]$.  $\|.\|$ denotes the $l_2$ norm for vectors and induced $l_2$ norm for matrices unless specified otherwise, and $'$ denotes transpose. We use $\M_\T$ to denote a sub-matrix of $\M$ formed by its columns indexed by entries in the set $\T$.
For a matrix $\P$ we use $\P^{(i)}$ to denote its $i$-th row.

A matrix $\P$ with mutually orthonormal columns is referred to as a {\em basis matrix} and is used to represent the subspace spanned by its columns.
For basis matrices $\P_1,\P_2$, we use $\SE(\P_1,\P_2):=\|(\I - \P_1 \P_1{}')\P_2\|$ as a measure of Subspace Error (distance) between their respective subspaces. This is equal to the sine of the largest principal angle between the subspaces. 
If $\P_1$ and $\P_2$ are of the same dimension, $\SE(\P_1, \P_2) = \SE(\P_2, \P_1)$.

We  use $\hat{\L}_{t; \alpha} := [\lhat_{t-\alpha + 1}, \cdots, \lhatt]$ to denote the matrix formed by $\lhat_t$ and $(\alpha-1)$ previous estimates.
Also, $r$-SVD$[\M]$ refers to the matrix of top $r$ left singular vectors of $\M$.%

A set $\Omega$ that is randomly sampled from a larger set (universe), $\mathcal{U}$, is said be {\em ``i.i.d. Bernoulli with parameter $\rho$''} if each entry of $\mathcal{U}$ has probability $\rho$ of being selected to belong to $\Omega$ independent of all others.

We reuse $C,c$ to denote different numerical constants in each use; $C$ is for constants greater than one and $c$ for those less than one. 

%

\begin{definition} [$\mu$-incoherence]
An $n \times r_{\scriptscriptstyle{P}}$ basis matrix $\P$ is $\mu$-incoherent if
$
\max_{i} \|\P^{(i)}\|_2^2 \le {\mu \frac{r_{\scriptscriptstyle{P}}}{n}}
$ ($\P^{(i)}$ is $i$-th row of $\P$). Clearly, $\mu \ge 1$.
%
\label{defmu}
\end{definition}

Throughout this paper, we assume that $f$, which is the condition number  of the population covariance of $\lt$, and the parameter, $\mu$,  are constants. This is assumed when the $\mathcal{O}(\cdot)$ notation is used.

\Subsection{Problem Statement}
ST-miss is precisely defined as follows.
At each time $t$, we observe a data vector $\yt \in \Re^n$ that satisfies%
\begin{align}
\yt = \proj(\lt) + \v_t, \text{ for } t = 1, 2, \dots, \tmax
\label{orpca_eq}
\end{align}
where $\proj(\z_i) = \z_i$ if $i \in \Omega_t$ and $0$ otherwise. Here $\v_t$ is small unstructured noise, $\Omega_t$ is the set of observed entries at time $t$, and $\lt$ is the true data vector that lies in a fixed or changing low ($r$) dimensional subspace of $\Re^n$, i.e., $\lt = \P_{(t)} \a_t$ where $\P_{(t)}$ is an $n \times r$ basis matrix with $r \ll n$. 
%
%
The goal is to track $\Span(\P_{(t)})$ and $\lt$ either immediately or within a short delay.
%
Denoting the set of missing entries at time $t$ as $\Tmisst$,  \eqref{orpca_eq} can also be written as
\begin{align}
\yt := \lt - \I_{\Tmisst} \I_{\Tmisst}{}' \lt + \vt.
\label{eq:dynmc}
\end{align}
We use $\z_t:=- \I_{\Tmisst}{}' \lt$ to denote the missing entries.
Clearly, $\Tmisst = (\Omega_t)^c$ (here $^c$ denotes the complement set w.r.t. $\{1,2,\dots,n\}$).
%
Writing $\yt$ as above allows us to tap into the  solution framework from earlier work \cite{rrpcp_perf,rrpcp_icml}. This was developed originally for solving robust ST which involves tracking $\lt$ and $\P_{(t)}$ from $\yt : = \lt + \vt + \xt$ where $\xt$ is a sparse vector with the outliers as its nonzero entries. ST-miss can be interpreted as its (simpler) special case if we let $\xt = - \I_{\Tmisst} \I_{\Tmisst}{}' \lt$. It is simpler because the support of $\xt$, $\Tmisst$, is known.

Defining the $n \times \tmax$ matrix $\L:= [\l_1, \l_2, \dots \l_d]$, the above is also a matrix completion (MC) problem; with the difference that for MC the estimates are needed only in the end (not on-the-fly). We use $\rmat$ to denote the rank of $\L$.

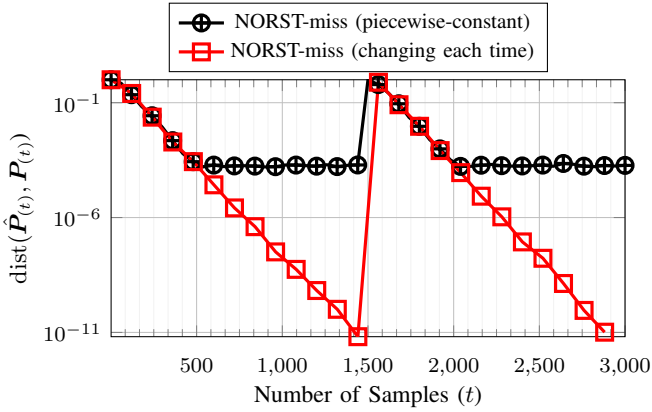
\begin{figure}[t!]
\centering
\begin{tikzpicture}
    \begin{groupplot}[
        group style={
            group size=1 by 1,
            horizontal sep=1cm,
            vertical sep=1cm,
                        x descriptions at=edge bottom,
           y descriptions at=edge left,
        },
        enlargelimits=false,
        width = .95\linewidth,
        height=5cm,
        grid=both,
    grid style={line width=.1pt, draw=gray!10},
    major grid style={line width=.2pt,draw=gray!50},
    minor tick num=5,
    ]
       \nextgroupplot[
		legend entries={
					NORST-miss (piecewise-constant),
					NORST-miss (changing each time),
					},
            legend style={at={(.85, 1.3)}},
            legend columns = 1,
            legend style={font=\footnotesize},
            ymode=log,
            xlabel={\small{Number of Samples ($t$)}},
            ylabel={\small{$\SE(\Phat_{(t)},\P_{(t)}$)}},
                                    xticklabel style= {font=\footnotesize, yshift=-1ex},
            yticklabel style= {font=\footnotesize},
        ]
                	        \addplot [black, line width=1.2pt, mark=oplus,mark size=3pt, mark repeat=2] table[x index = {0}, y index = {1}]{\pwjust};
	        \addplot [red, line width=1.2pt, mark=square,mark size=3pt, mark repeat=1, select coords between index={0}{24}] table[x index = {2}, y index = {3}]
	        {\pwjust};
	        \end{groupplot}
\end{tikzpicture}
\caption{\small{ 
Demonstrating the need for the piecewise constant subspace change model. The black circles plot is for subspace changing at each time $t$, while the red squares one is for piecewise constant subspace change, with change occurring at $t=t_1$. The data is generated so that, in both experiments, $\SE(\P_{(t_1)}, \P_{(0)})$ is the same. In the piecewise constant case (red squares), we can achieve near perfect subspace recovery. But this is not possible in the ``changing at each time'' (black circles) case. 
For details, see Sec. \ref{sec:sims_main} and Fig. \ref{fig:stmiss}(c).
}}
\label{fig:pwconst}
\vspace{.2cm}
\end{figure}

\Subsection{Identifiability assumptions} \label{identif}
The above problem definition does not ensure identifiability. If $\L$ is sparse, it is impossible to recover it from a subset of its entries. Moreover, even if it is dense, it is impossible to complete it if all the missing entries are from a few rows or columns.
Finally, if the subspace changes at every time $t$, the number of unknowns ($nr$) is more than the amount of available data at time $t$ ($n$) making it impossible to recover all of them.

One way to ensure subspaces' identifiability is to assume that they are piecewise constant with time, i.e., that
\[
\P_{(t)} = \P_j \text{ for all } t \in [t_j, t_{j+1}), \ j=1,2,\dots, J.
\]
with $t_{j+1}-t_j \ge r$. Let $t_0=1$ and $t_{J+1}=\tmax$.  This ensures that at least $r$ $n$-dimensional data vectors $\yt$ are available (this is the minimum needed to compute the subspace even if perfect data $\yt=\lt$ were available). The $t_j$'s are the subspace change times.  With this model,  $\rmat \le r J$.
{
 When the above model is not assumed, one cannot track to any desired accuracy, see the black circles plot in Fig. \ref{fig:pwconst}. This is because the subspace change at each time can be interpreted as a $r$-dimensional piecewise constant subspace change plus noise. To understand this precisely, consider the first $\alpha$ frames, for any $\alpha \ge r$. Let $\P$ be the matrix of top $r$ left singular vectors of $[\P_{(0)}, \P_{(1)}, \dots, \P_{(\alpha-1)}]$. Then, in this interval, $\yt := \mathcal{P}_{\Omega_t}( \P_{(t)} \at )$ can be rewritten as $\yt = \mathcal{P}_{\Omega_t}( \P (\P'\P_{(t)} \at) ) + \vt$ where $\vt =  \mathcal{P}_{\Omega_t}(\P_{(t)} \at -  \P (\P'\P_{(t)}) \at )$. A similar argument can be extended to any set of $\alpha$ frames.
}

As explained in earlier work on MC \cite{matcomp_first, matcomp_candes, recht_mc_simple}, one way to ensure that $\L$ is not sparse is to assume that its left and right singular vectors are dense. This is the well-known incoherence or denseness assumption. Left singular vectors incoherent is nearly equivalent to imposing $\mu$-incoherence of the $\P_j$'s with $\mu$ being a numerical constant. As explained in \cite[Remark 2.4]{rrpcp_icml}, the following assumption on $\at$'s is similar to right incoherence, and hence we call it ``statistical right incoherence''.%
\begin{definition}[Statistical Right Incoherence]
We assume that the $\at$'s are zero mean, i.e., $\E[\at]= 0$; are mutually independent over time; have identical diagonal covariance matrix $\Lam$, i.e., that $\E[\at \at{}'] = \Lam$ with $\Lam$ diagonal; and are element-wise bounded. 
Element-wise bounded means that  there exists a numerical constant $\mu \ge 1$, such that $\max_i \max_t (\a_t)_i^2 \le \mu \max_t \lambda_{\max}(\E[\a_t\a_t{}'])$. This  implies that the $\a_t$'s are sub-Gaussian with sub-Gaussian norm bounded by $\mu \max_t \lambda_{\max}(\E[\a_t\a_t{}']) = \mu \lambda_{\max}(\Lam)$. A simple example of element-wise bounded random vectors (r.v) is uniform r.v.s. 
\label{def_right_incoh}
\end{definition}


Motivated by the Robust PCA literature \cite{robpca_nonconvex}, one way to ensure that the missing entries are spread out is to bound the maximum fraction of missing entries in any row and in any column. We use $\missfracrow$ and $\missfraccol$ to denote these. Since NORST-miss is a mini-batch approach that works on batches of $\alpha$ frames, we actually need to bound the maximum fraction of missing entries in any sub-matrix of $\L$ with $\alpha$ consecutive columns. We denote this by $\missfracrow_{\alpha}$. We precisely define these below.
%
{
\begin{definition}[$\missfraccol, \missfracrow_\alpha$]
For a  discrete time interval, $\J$, let
$
\gamma(\J): = \max_{i=1,2,\dots,n} \frac{1}{|\J|} \sum_{t \in \J} \one_{ \{i \in \Tmisst \} }
$
where $\one_{S}$ is the indicator function for statement $S$. Thus, $\sum_{t \in \J} \one_{ \{ i \in \Tmisst \} }$ counts the maximum number of missing entries in row $i$ of the sub-matix $\L_\J$ of the data matrix $\L:=[\l_1, \l_2, \dots,\l_d]$. So, $\gamma(\J)$ is the maximum fraction of missing entries in any row of $\L_\J$.
Let $\J^\alpha$ denote a time interval of duration $\alpha$. Then,
$
\missfracrow_{\alpha}:= \max_{\J^\alpha \subseteq [1, \tmax]} \gamma(\J^\alpha).
$
Also, $\missfraccol :=  \max_t|\Tmisst|/n$.
\end{definition}
}

\section{The NORST-miss algorithm and guarantees} \label{sec:norstmiss}
We explain the basic algorithm next. We give and discuss the guarantee for the noise-free $\vt=0$ case in Sec. \ref{sec:mainres_noisefree}. The corollary for the noisy case is given in Sec. \ref{sec:mainres_noisy}. Extensions of basic NORST-miss are given in Sec. \ref{sec:ext}.

\Subsection{NORST-miss algorithm} \label{sec: norstmiss_algo}


The complete psedo-code for our algorithm is provided in Algorithm \ref{algo:NORST-st-basic}. After initialization, the algorithm iterates between a {\em projected Least Squares (LS)} step and a {\em Subspace Update (including Change Detect)} step. Broadly, projected LS estimates the missing entries of $\lt$ at each time $t$. Subspace update toggles between the ``update'' phase and the change ``detect'' phase. In the update phase, it improves the estimate of the current subspace using a short mini-batch of ``filled in'' versions of $\lt$. In the detect phase, it uses these to detect subspace change. 

\textbf{Initialization:}
The algorithm starts in the ``update'' phase and with zero initialization: $\Phat_0 \leftarrow \bm{0}_{n \times r}$. For the first $\alpha$ frames, the projected LS step (explained below) simply returns $\lhat_t = \yt$. Thus, a simpler way to understand the initialization is as follows: wait until $t=\alpha$ and then compute the first estimate of $\Span(\P_0)$ as the $r$-SVD (matrix of top $r$ left singular vectors) of $[\y_1, \y_2, \dots \y_\alpha]$. This step is solving a PCA with missing data problem which, as explained in \cite{pca_dd_isit}, can be interpreted as a problem of PCA in sparse data-dependent noise. Because we assume that the number of missing entries at any time $t$ is small enough, and the set of missing entries changes sufficiently over time\footnote{Equivalently, we bound the maximum number of missing entries in any column and in any row of the data matrix}, we can prove that this step gives a good first estimate of the subspace. 


%

\textbf{Projected LS:}
Recall that NORST-miss is a modification of  NORST for robust ST from \cite{rrpcp_icml}. In robust ST,  sudden subspace changes cannot be detected because these are confused for outliers. Its projected-LS step is thus deigned using a slow (small) subspace change assumption. However, as we will explain later, for the current missing data setting, it also works in case of sudden changes.
Suppose that the previous subspace estimate, $\Phat_{(t-1)}$, is a ``good enough'' estimate of the previous subspace $\P_{(t-1)}$.
Under slow subspace change, it is valid to assume that $\Span(\P_{(t-1)})$ is either equal to or close to $\Span(\P_{(t)})$. Thus, under this assumption, it is a good idea to project $\yt$ onto the orthogonal complement of $\Phat_{(t-1)}$ because this will nullify most of $\lt$, i.e., the not-nullified part of $\lt$, $\bt: =  \bpsi \lt$, will be small. Here $\bpsi := \I - \Phat_{(t-1)}\Phat_{(t-1)}{}'$.
Using this idea, we compute $\tty_t:= \bpsi  \yt = \bpsi _{\Tmisst} \zt + \bt + \bpsi \vt$. 
Estimating $\z_t$ can be interpreted as a LS problem $\min_{\bm{z}} \|\tty_t - \bpsi_{\Tmisst} \bm{z}\|^2$. Solving this gives
\begin{align}
\hat{\bm{z}}_t = \left(\bpsi_{\Tmisst}{}'\bpsi_{\Tmisst} \right)^{-1} \bpsi_{\Tmisst}{}'\tty_t.
\label{eq:zhatt}
\end{align}
Next, we use to this to compute $\lhat_t = \yt - \I_{\Tmisst} \hat{\bm{z}}_t$. Observe that the missing entries $\z_t$ are recoverable as long as $\bpsi_{\Tmisst}$ is well-conditioned. A necessary condition for this is $(n-r) > |\Tmisst|$. As we will see later, a sufficient condition is $|\Tmisst| < c n / r$ because this ensures that the restricted isometry constant (RIC) \cite{candes_rip} of $\bpsi$ of level $|\Tmisst|$ is small.

In settings where $\Span(\P_{(t-1)})$ is {\em not} close to $\Span(\P_{(t)})$ (sudden subspace change), the above approach still works.
Of course, in this case, it is not any better (or worse) than re-initialization to zero, because, in this case, $ \|\bpsi \lt\|$ is of the same order as $\|\lt\|$. We can use the same arguments as those used for the initialization step to argue that the first subspace update works even in this case.





\textbf{Subspace Update:}
The $\lhatt$'s are used for subspace update. In its simplest (and provably correct) form, this is done once every $\alpha$ frames by $r$-SVD on the matrix formed by the previous $\alpha$ $\lhat_t$'s. 
Let $\that_j$ be the time at which the $j$-th subspace change is detected (let $\that_0 := 0$). For each $k=1,2,\dots, K$, at $t = \that_j + k \alpha-1$, we compute the $r$-SVD of $\Lhat_{t; \alpha}$ to get $\Phat_{j,k}$ ($k$-th estimate of subspace $\P_j$).  After $K$ such updates, i.e., at  $t =  \that_j + K\alpha - 1:=\that_{j,fin}$ the update is complete and the algorithm enters the ``detect'' phase.
Each update step is a PCA in sparse data-dependent noise problem. This allows us to use the result from \cite{pca_dd_isit} to show that, as long as the missing entries' set changes enough over time ($\missfracrow_\alpha$ is bounded for each interval), each update step reduces the subspace recovery error to $0.3$ times its previous value. Thus, by setting $K=C \log (1/\zz)$, one can show that, after $K$ updates, the subspace is recovered to $\zz$ accuracy.

\textbf{Subspace change detect:}
To simply understand the detection strategy, assume that the previous subspace $\P_{j-1}$ has been estimated to $\zz$ accuracy by $t= \that_{j-1,fin} = \that_{j-1} + K\alpha - 1$ and denote it by $\Phat_{j-1}:= \Phat_{j-1,K}$. Also assume that $\vt=0$.
At every $t = \that_{j-1,fin} + u \alpha-1$, $u=1,2,\dots$, we detect change by checking if the maximum singular value of the matrix $(\I-\Phat_{j-1}\Phat_{j-1}{}')  \Lhat_{t;\alpha}$ is above a pre-set threshold, $\sqrt{\lthres \alpha}$, or not. This works because, if the subspace has not changed, this matrix will have all singular values of order $\zz \sqrt{\lambda^+}$. If it has changed, its largest singular value will be at least $\SE(\P_{j-1}, \P_j) \sqrt{\lambda^-}$. By picking $\zz$ small enough, one can ensure that, whp, all changes are detected.

\textbf{NORST-miss-smoothing for MC:}
The above is the tracking/online/filtering mode of NORST-miss. It outputs an estimate of $\lt$ as soon as a new measurement vector $\yt$ arrives and an estimate of the subspace every $\alpha$ frames. Notice  that, order-wise, $\alpha$ is only a little more than $r$ which is the minimum delay needed to compute the subspace even if perfect data $\yt=\lt$ were available.
Once an $\zz$-accurate estimate of the current subspace is available, one can improve all past estimates of $\lt$ to ensure that all estimates are $\zz$-accurate. This is called the smoothing mode of operation. To be precise, this is done as given in line $25$ of Algorithm \ref{algo:NORST-st-basic}. This allows us to get a completed matrix $\Lhat$ with all columns being $\zz$-accurate.

\textbf{Memory Complexity:}
In online or filtering mode, NORST-miss needs $\alpha =O( r \log n)$ frames of storage. In smoothing mode, it needs $\mathcal{O}((K+2)\alpha)=\mathcal{O}( r \log n \log(1/\epsilon))$ frames of memory. Therefore its memory complexity, even in the smoothing mode, is just  $\mathcal{O}(n r \log n \log(1/\epsilon))$. Thus, it provides a nearly memory-optimal mini-batch solution for MC.

\textbf{Algorithm parameters:}
The algorithm has 4 parameters: $r$, $K$, $\alpha$, and $\lthres$. Theoretically these are set as follows: assume  that $r,\lambda^+,\lambda^-$ are known and pick a desired recovery error $\zz$. Set $\alpha = C_1 f^2 r \log n$ with $f=\lambda^+/\lambda^-$, $K=C_2 \log(1/\zz)$ and $\lthres = c \lambda^-$ with $c$ a small constant. We explain practical approaches in Sec \ref{sec:sims_main}.%

\newcommand{\xmin}{x_{\min}}
\newcommand{\rrow}{\rho_{\text{row}}}
\newcommand{\rcol}{\rho_{\text{col}}}

\Subsection{Main Result: noise-free ST-miss and MC} \label{sec:mainres_noisefree}
First, for simplicity, consider the noise-free case, i.e., assume $\vt = 0$.
Let $\dif_j:= \SE(\P_{j-1}, \P_j)$.

\begin{theorem}[NORST-miss, $\vt=0$ case]
Consider Algorithm \ref{algo:NORST-st-basic}. Let $\alpha := C f^2  r \log n$,  $\Lam:= \E[\a_1 \a_1{}']$, $\lambda^+:=\lambda_{\max}(\Lam)$,  $\lambda^-:=\lambda_{\min}(\Lam)$, $f:=\lambda^+/\lambda^-$.

Pick an $\zz \leq \min(0.01,0.03 \min_j \SE(\P_{j-1}, \P_j)^2/f)$. Let $K := C \log (1/\zz)$.
If
\ben
\item  left and statistical right incoherence: $\P_j$'s are $\mu$-incoherent and $\at$'s satisfy statistical right incoherence (Definition \ref{def_right_incoh});

\item $\missfraccol \le  \frac{c_1}{\mu r}$, $\missfracrow_{\alpha} \le \frac{c_2}{f^2}$;

\item subspace change: assume $t_{j+1}-t_j > C r \log n \log(1/\zz)$;

\item $\at$'s are independent of the set of missing entries $\Tmisst$;


\een
then, with probability (w.p.) at least $1 - 10 \tmax n^{-10} $,
\ben
\item subspace change is detected quickly: $t_j \le \that_j \le t_j+2 \alpha$,
\item the subspace recovery error satisfies
\[
\SE(\Phat_{(t)}, \P_{(t)}) \le
 \left\{
\begin{array}{ll}
(\zz + \dif_j) & \text{ if }  t \in \J_1, \\
 (0.3)^{k-1} (\zz + \dif_j) & \text{ if }  t \in \J_k, \\
\zz   & \text{ if }  t \in \J_{K+1}.
\end{array}
\right.
\]
\item and {$\|\lhat_t-\lt\| \le 1.2 (\SE(\Phat_{(t)}, \P_{(t)}) + \zz) \|\lt\|$}. 
\een
Here $\J_0:= [t_j, \that_j+\alpha)$, $\J_{k} := [\that_j+k\alpha, \that_j+ (k+1)\alpha)$ and $\J_{K+1} := [\that_j+ (K+1)\alpha, t_{j+1})$ and $\dif_j:= \SE(\P_{j-1}, \P_j)$.

The memory complexity is $\mathcal{O}(n r \log n \log(1/\zz))$ and the time complexity is $\mathcal{O}(n \tmax r \log(1/\zz))$.
\label{thm:stmiss}
\end{theorem}

\begin{corollary}[NORST-miss for MC]\label{cor:thm1}
Under the assumptions of Theorem \ref{thm:stmiss}, NORST-miss-smoothing (line 25 of Algorithm \ref{algo:NORST-st-basic}) satisfies  {$\|\lhat_t-\lt\| \le \zz \|\lt\|$ } for all $t$.
Thus, $\|\Lhat - \L\|_F     \le \zz \|\L\|_F$.
\end{corollary}
The proof is similar to that given in \cite{rrpcp_icml} for the correctness of NORST for robust ST. Please see the Appendix for the changes.

{
For the purpose of this discussion, we treat the condition number $f$ and the incoherence parameter $\mu$ as constants.
The above result proves that NORST-miss tracks piecewise constant subspaces to $\epsilon$ accuracy, within a delay that is near-optimal, 
 under the following assumptions: left and ``statistical'' right incoherence holds; the fraction of missing entries in any column of $\L$ is $\mathcal{O}(1/r)$ while that in any row (of $\alpha$-consecutive column sub-matrices of it) is $\mathcal{O}(1)$. Moreover, ``smoothing mode'' NORST-miss returns $\zz$-accurate estimates of each $\lt$ and thus also solves the MC problem. Even in this mode, it  has  {\em near-optimal memory complexity and is order-wise as fast as vanilla PCA}. The above result is the {\em first complete guarantee} for ST-miss. Also, unlike past work, it can {\em deal with piecewise constant subspaces while also automatically reliably detecting subspace change} with a near-optimal delay.

Consider the total number of times a subspace can change, $J$. Since we need the subspace to be constant for at least $(K+3)\alpha$ frames, $J$ needs to satisfy $J (K+3) \alpha \le d$. Since we need $(K+3)\alpha$ to be at least $C r \log n \log(1/\zz)$, this means that $J$ must satisfy
\[
J \le c \frac{d}{r \log n \log(1/\zz)}.
\]
This, in turn, implies that the rank of the entire matrix, $\rmat$, can be at most`
\[
\rmat = r J \le  c \frac{d}{\log n \log (1/\zz)}.
\]
Observe that this upper bound is nearly linear in $d$. This is what makes our corollary for MC interesting.
It implies that we can recover $\L$ to $\zz$ accuracy even in this {\em nearly linearly growing rank regime}, of course only if the subspace changes are piecewise constant with time and frequent enough so that $J$ is close to its upper bound.
In contrast, existing MC guarantees, these require left and right incoherence of $\L$ and a Bernoulli model on observed entries with observation probability $m / n d$ where $m$ is the required number of observed entries on average. The convex solution \cite{recht_mc_simple} needs $m = C n \rmat \log^2 n$ while the best non-convex solution \cite{rmc_gd} needs $m = C n \rmat^2 \log^2 n $ observed entries. The non-convex approach is much faster, but its required $m$ depends on $\rmat^2$ instead of $\rmat$ in the convex case. See Sec. \ref{sec:prior_art} for a detailed discussion, and Table \ref{tab:comp_mc} for a summary of it.
On the other hand, our missing fraction bounds imply that the total number missing entries needs to at most $ \min( n d \cdot \missfracrow, d n \cdot  \missfraccol) = c \frac{nd}{r}$, or that we need at least $m = (1 - c/r) nd$ observed entries.

If subspace changes are infrequent ($J$ is small) so that $\rmat \approx r \ll d$, our requirement on observed entries is much stronger than what existing MC approaches need. 
However, suppose that $J$ equals its allowed upper bound so that $\rmat =  c \frac{d}{\log n \log (1/\zz)}$; but $r$ is small, say $r = \log n$. In this setting, we need $nd(1 - c/\log n)$ while the convex MC solution needs $c n  \frac{d}{\log n \log (1/\zz)} \log^2 n  = c n d \frac{\log n}{\log(1/\zz)}$. If $\zz =1/n$, this is $c \cdot nd$, if $\zz$ is larger, this is even larger than $c \cdot nd$. Thus, in this regime, our requirement on $m$ is only a little more stringent. Our advantage is that we do not require a Bernoulli (or any probability) model on the observed entries' set {\em and} our approach is much faster, memory-efficient, and nearly delay-optimal. This is true both theoretically and in practice; see Tables \ref{tab:comp_mc} and \ref{tab:all_MCalgos_frob}. If we consider non-convex MC solutions, they are much faster, but they cannot work in this nearly linear rank regime at all because they will need $C n d^2 / \log^2 n$ observed entries, which is not possible.


A possible counter-argument to the above can be: what if one feeds smaller batches of data to an MC algorithm. Since the subspace change times are not known, it is not clear how to do this. One could feed in batches of size $K \alpha$ which is the memory size used by NORST-miss-smoothing. Even in this case the discussion is the same as above. To simplify writing suppose that $\zz = 1/n$. The convex solution will need  $m = c n (C r \log^2 n )$ observed entries for a matrix of size $n \times (C r \log^2 n)$. Thus $m$ required is again linear in the matrix size. NORST-miss-smoothing will need this number to be $ (1 - c/r) n (C r \log^2 n) $ which is again only slightly worse when $r$ is small. The non-convex methods will again not work. 

The Bernoulli model on the observed entries' set can often be an impractical requirement. For example, erasures due to transmission errors or image/video degradation often come in bursts. Similarly video occlusions by foreground objects are often slow moving or occasionally static, rather than being totally random.
Our guarantee does not require the Bernoulli model but the tradeoff is that, in general, it needs more observed entries.
A similar tradeoff is observed in the robust PCA literature. The guarantee of \cite{rpca} required a uniform random or Bernoulli model on the outlier supports, but tolerated a constant fraction of corrupted entries. In other words it needed the number of uncorrupted entries to be at least $ c \cdot nd$. Later algorithms such as AltProj \cite{robpca_nonconvex} did not require any random model on outlier support but needed the number of un-corrupted entries to be at least $(1 -c/r) nd$ which is a little more stringent requirement.

\begin{table}[t!]
{
\caption{\small{List of Symbols and Assumptions used in Theorem \ref{thm:stmiss}.}}
\vspace{.1cm}
\centering
\resizebox{\linewidth}{!}{
\renewcommand{\arraystretch}{1.5}
\begin{tabular}{cccc} \toprule
\multicolumn{2}{c}{{\bf Observations: } $\yt = \proj(\lt) + \vt = \proj(\P_{(t)} \at) + \vt$} \\ \toprule
Symbol & Meaning \\ \midrule
$t_j$ & $j$-th subspace change time \\
for $t \in [t_j, t_{j+1})$, $\P_{(t)} = \P_j$ &  Subspace at time $t$ \\

$\proj(\cdot)$ & mask to select elements present in $\Omega_t$ \\
$\Omega_t$ & Support set of observed entries \\
$\Tmisst (= \Omega_t^c) $ & Support set of missing entries  \\
$\vt$ & dense, unstructured noise \\
\midrule
\multicolumn{2}{c}{{\bf Principal Subspace Coefficients} ($\at$'s)} \\
\midrule
\multicolumn{2}{c}{element-wise bounded, zero mean,} \\
\multicolumn{2}{c}{mutually independent with identical and diagonal covariance}  \\
\multicolumn{2}{c}{$\ep{[\at \at{}']} := \Lam$}  \\
$\lambda_{\max}(\Lam) = \lambda^+ (\lambda_{\min}(\Lam) = \lambda^-)$ & Max. (min.) eigenvalue of $\Lam$ \\
$f := \lambda^+/\lambda^-$ & Condition Number of $\Lam$ \\
\midrule
\multicolumn{2}{c}{{\bf Missing Entries} ($\zt = - \I_{\Tmisst}{}' \lt$)} \\ \midrule
Row-Missing Entries & $\missfracrow_{\alpha} \leq  0.001/f^2$  \\
Column-Missing Entries & $\missfraccol \leq 0.01/\mu r$   \\ \midrule
\multicolumn{2}{c}{{\bf Intervals for $j$-th subspace change and tracking}} \\ \midrule
$\that_j$ & $j$-th subspace change detection time \\
$\that_{j, fin}$ & $j$-th subspace update complete \\
$\J_0 := [t_j, \that_j)$ & interval before $j$-th subspace change detected \\
$\J_k := [\that_j + (k-1)\alpha, \that_j + k\alpha)$ & $k$-th subspace update interval \\
$\J_{K+1}:=[\that_j + K\alpha, t_{j+1})$ &  subspace update completed
\\ \midrule
\multicolumn{2}{c}{{\bf Algorithm \ref{algo:NORST-st-basic} Parameters}} \\ \midrule
$\alpha$ & \# frames used for subspace update \\
$K$ & \# of subspace updates for each $j$ \\
$\lthres$ & threshold for subspace detection \\
\bottomrule
\end{tabular}
\label{tab:not1}}
}
\end{table}

\Subsection{Main Result -- ST-miss and MC with noise} \label{sec:mainres_noisy}
So far we gave a result for ST-miss and MC in the noise-free case. A more practical model is one that allows for small unstructured noise (modeling error). Our result also extends to this case with one extra assumption.
In the noise-free case, there is no real lower bound on the amount of subspace change required for reliable detection. Any nonzero subspace change can be detected (and hence tracked) as long as the previous subspace is recovered to $\zz$ accuracy with $\zz$ small enough compared to the amount of change. 
If the noise $\vt$ is such that its maximum covariance in any direction is smaller than $\zz^2 \lambda^-$, then Theorem \ref{thm:stmiss} and Corollary \ref{cor:thm1} hold with almost no changes. If the noise is larger, as we will explain next, we will need the amount of subspace change to be larger than the noise-level. Also, we will be able to track the subspaces only up to accuracy equal to the noise level.

Suppose that the noise $\vt$ is bounded. Let $\lambda_v^+:=\|\E[\vt \vt{}']\|$ be the noise power and let $r_v:= \max_t \|\vt\|^2 / \lambda_v^+$ be the effective noise dimension. Trivially, $r_v \le n$. To understand things simply, first suppose that the subspace is fixed. If the noise is isotropic (noise covariance is a multiple of identity), then, as correctly pointed out by an anonymous reviewer, one can achieve noise-averaging in the PCA step by picking $\alpha$ large enough: it needs to grow as \footnote{$\alpha$ needs to grow as $ C \min(r_v \log n, n) (\lambda_v^+/\lambda^-)/\epsilon^2$; for the isotropic case, $r_v = n$ and thus the discussion follows.} $ n (\lambda_v^+ /\lambda^-)/\zz^2$. Isotropic noise is the most commonly studied setting for PCA, but it is not the most practical. In the more practical non-isotropic noise case, it is not even possible to achieve noise-averaging by increasing $\alpha$. In this setting, with any choice of $\alpha$, the subspace can be recovered only up to  the noise level, i.e., we can only achieve recovery accuracy $c \lambda_v^+ /\lambda^-$. If we are satisfied with slightly less accurate estimates, i.e., if we set $\zz= c \sqrt{\frac{\lambda_v^+}{\lambda^-} }$, and if the effective noise dimension $r_v = C r$, then the required value of $\alpha$ does not change from what it is in Theorem \ref{thm:stmiss}.
%
Now consider the changing subspace setting.  We can still show that we can detect subspace changes that satisfy $0.03 \min_j \SE(\P_{j-1}, \P_j)^2/f \ge \zz$, but now $\zz =  c \sqrt{\frac{\lambda_v^+}{\lambda^-} }$.  This imposes a non-trivial lower bound on the amount of change that can be detected. The above discussion is summarized in the following corollary.

\begin{corollary}[ST-miss and MC with $\vt \neq 0$]\label{cor:noisy}
Suppose that $\vt$ is  bounded, mutually independent and identically distributed (iid) over time, and is independent of the $\lt$'s. Define $\lambda_v^+:=\|\E[\vt \vt{}']\|$ and $r_v:= \frac{\max_t\|\vt\|^2}{\lambda_v^+}$.
\bi
\item If  $r_v = C r$ and $\lambda_v^+ \le c \zz^2 \lambda^-$, then the results of Theorem \ref{thm:stmiss} and Corollary \ref{cor:thm1} hold without any changes.

\item For a general $\lambda_v^+$, we have the following modified result. Suppose that  $r_v = C r$, $\min_j \SE(\P_{j-1}, \P_j)^2 \ge C f \sqrt{\frac{\lambda_v^+}{\lambda^-} }$, and conditions 1, 2, 3 of Theorem \ref{thm:stmiss} hold. Then all conclusions of Theorem \ref{thm:stmiss} and Corollary \ref{cor:thm1} hold with $\zz = c \sqrt{\frac{\lambda_v^+}{\lambda^-} }$.

\item For a general $r_v $, if we set $\alpha = C f^2 \max(r \log n, \min(n, r_v \log n))$ then the above conclusions hold.
\ei
\end{corollary}

If the noise is {\em isotropic}, the next corollary shows that we can track to any accuracy $\zz$ by increasing the value of $\alpha$. It is not interesting from a tracking perspective because its required value of $\alpha$ is much larger. However, it provides a result that is comparable to the result for streaming PCA with missing data from \cite{streamingpca_miss} that we discuss later.

\begin{corollary}[ST-miss and MC, isotropic noise case]\label{cor:noisy2}
If the noise $\vt$ is isotropic (so that $r_v=n$), then, for any desired recovery error level $\zz$,
if $\alpha = C  n \frac{ \frac{\lambda_v^+}{\lambda^-} }{\zz^2} $, and all other conditions of Theorem \ref{thm:stmiss} hold, then all conclusions of Theorem \ref{thm:stmiss} and Corollary \ref{cor:thm1} hold.
\end{corollary}

We should mention here that the above discussion and results assume that PCA is solved via a simple SVD step (compute  top $r$ left singular vectors). In the non-isotropic noise case, if its covariance matrix were known (or could be estimated), then one can replace simple SVD by pre-whitening techniques followed by SVD,  in order to get results similar to the isotropic noise case, e.g., see \cite{leeb_non_iso}.

\renewcommand{\arraystretch}{1.3}
\begin{table*}[t!]
{
    \caption{\small{Comparing guarantees for ST-miss. We treat the condition number and incoherence parameters as constants for this discussion. 
    }}
\begin{center}
        \resizebox{.9\linewidth}{!}{
    \begin{tabular}{ccccccc}
        \toprule
        \textbf{Algorithm} &\textbf{Tracking} & \textbf{Memory} & \textbf{Time} & \textbf{Allows changing} & \textbf{Observed Entries} \\
                &\textbf{delay} & &  & \textbf{subspaces?}  \\
        \midrule
GROUSE \cite{grouse} & Partial Guarantee & $\mathcal{O}(nr)$ & $\mathcal{O}(n d \fracobs r^2)$ & No & i.i.d. Bernoulli($\fracobs$) \\
PETRELS \cite{petrels_new} & Partial Guarantee & $\mathcal{O}(nr^2)$ & $\mathcal{O}(n d \fracobs r^2)$ & No & i.i.d. Bernoulli($\fracobs$) \\
MBPM \cite{streamingpca_miss, eldar_jmlr_ss} & $  d \succsim \frac{r^2 \log^2 n \log (1/\zz)}{\fracobs^2}$ & $\mathcal{O}(nr)$ & $\mathcal{O}(ndr)$    & No & i.i.d. Bernoulli($\fracobs$)    \\
                                            NORST-miss & $d \geq r \log n \log(1/\epsilon)$ & $\mathcal{O}\left(nr \log n \log \frac{1}{\epsilon}\right)$ & $\mathcal{O}\left(n d r \log \frac{1}{\epsilon}\right)$ & Yes & bounded fraction,  \\
(this work) & & & & & $c/r$ per column, $c$ per row \\
    \bottomrule
    \end{tabular}
    }
    \label{tab:comp_st}
    \end{center}
}
%
\caption{\small{Comparing MC guarantees. 
Recall $\rmat:=\rank(\L) \le r J$.  In the regime when the subspace changes frequently so that $J$ equals its upper bound and $\rmat \approx d/\log^2 n$, NORST-miss is better than the non-convex methods (AltMin, projGD, SGD) and only slightly worse than the convex ones (NNM). In general, the sample complexity for NORST-miss is significantly worse than all the MC methods.
}}
\vspace{.14cm}
\begin{center}
		\resizebox{.95\linewidth}{!}{
	\begin{tabular}{@{}c@{}c@{}c@{}c@{}c@{}c@{}c}
		\toprule
		\textbf{Algorithm} & \textbf{Sample complexity} & \textbf{Memory} & \textbf{Time} & \multicolumn{2}{c}{\textbf{Observed entries}} \\ 
		& \textbf{(\# obs. entries, $m$)} & \ & \ & \multicolumn{2}{c}{} \\ 
		\midrule
		nuc norm min (NNM) \cite{matcomp_first} & $\Omega(n \rmat \log^2 n)$ & $\mathcal{O}(nd)$ & $\mathcal{O}(n^3/\sqrt{\epsilon})$ & \multicolumn{2}{c}{i.i.d. Bernoulli ($m/nd$)} \\
		
						weighted NNM \cite{coherent_mc} & $\Omega(n \rmat \log^2 n)$ & $\mathcal{O}(nd)$  & $\mathcal{O}(n^3 /\sqrt{\epsilon})$
		& \multicolumn{2}{c}{indep. Bernoulli} \\  

		AltMin \cite{optspace} & $\Omega(n \rmat^{4.5}  \log \frac{1}{\epsilon})$ & $\mathcal{O}(nd)$ & $\mathcal{O}(n \rmat \log \frac{1}{\epsilon})$
		& \multicolumn{2}{c}{i.i.d. Bernoulli ($m/nd$)} \\
		projected-GD \cite{rmc_gd} & $\Omega(n \rmat^2 \log^2 n)$ & $\mathcal{O}(nd)$  & $\mathcal{O}(n \rmat^3 \log^2 n \log \frac{1}{\epsilon})$
		& \multicolumn{2}{c}{i.i.d. Bernoulli ($m/nd$)} \\

online SGD \cite{onlineMC1} & $\Omega\left(n \rmat^2 \log n \left(\rmat+\log \frac{1}{\epsilon}\right)\right)$ & $\mathcal{O}(nd)$ & $\mathcal{O}\left(n \rmat^4 \log n  \log \frac{1}{\epsilon}\right)$ & \multicolumn{2}{c}{i.i.d. Bernoulli ($m/nd$)} \\
\textbf{NORST-miss} & $\bm{\Omega((1 - \frac{c}{r}) nd)}$ & $\mathcal{O}\left(nr \log n \log \frac{1}{\epsilon}\right)$ & $\mathcal{O}\left(n d r \log \frac{1}{\epsilon}\right)$ & \multicolumn{2}{c}{\textbf{$\le c \cdot d$ per row}} \\
{\bf (this work)} & &  & & \multicolumn{2}{c}{\textbf{$\le (1 -\frac{c}{r}) \cdot n$ per column}}  \\

{\bf Sample-Efficient} & $\Omega (  n \rmat \log^2 n  \log r )$ \ & \ $\mathcal{O}\left(nr \log n \log \frac{1}{\epsilon}\right)$ & $\mathcal{O}\left(n d r \log \frac{1}{\epsilon}\right)$ &  \multicolumn{2}{c}{i.i.d. Bernoulli($\rho_t$) where,} \\
{\bf NORST-miss} & & & & \multicolumn{2}{c}{$\rho_t = 1-c/r$ for $t \in [t_j, t_j +(K+2)\alpha)$} \\
{\bf (this work)}  & & & & \multicolumn{2}{c}{$\rho_t = r \log^2 n \log r  /nd$ other times} \\
	\bottomrule
	\end{tabular}
	}
	\label{tab:comp_mc}
	\end{center}
	{\em Note:} 	Here, $f(n) = \Omega(g(n))$ implies that there exists a $G>0$ and an $n_0 > 0 $ s.t for all $n > n_0, \  |f(n)| \geq G \cdot |g(n)|$
	\vspace{-.2in} 

\end{table*}

\begin{algorithm}[t!]
\caption{NORST-miss.}
\label{algo:NORST-st-basic}
\begin{algorithmic}[1]
\STATE \textbf{Input}:  $\yt$, $\Tmisst$  \textbf{Output}:  $\lhatt$,  $\Phat_{(t)}$ \textbf{Parameters:} $r$, $K=~C\log(1/\zz)$, $\alpha= C f^2 r \log n$, $\lthres =2 \zz^2 \lambda^+$
\STATE $\Phat_0 \leftarrow \bm{0}_{n \times r}$, $\tildej~\leftarrow~1$, $k~\leftarrow~1$
\STATE
$\mathrm{phase} \leftarrow \mathrm{update}$; $\that_{0} \leftarrow 0$; $\that_{-1, fin} = 0$
\FOR {$t > 0$}
\STATE $\bpsi \leftarrow \bm{I} - \hat{\pt}_{(t-1)}\hat{\pt}_{(t-1)}{}'$; $\tty_t \leftarrow \bpsi \yt$;
\STATE  $\hat{\bm{\ell}}_t \leftarrow \yt - \I_{\Tmisst} ( \bpsi_{\Tmisst}{}' \bpsi_{\Tmisst} )^{-1} \bpsi_{\Tmisst}{}'\tty_t$.

\IF{$\text{phase} = \text{update}$}
\IF {$t = \that_j + u \alpha - 1$ for $u = 1,\ 2,\ \cdots,$}
\STATE $\Phat_{j, k} \leftarrow  r$-SVD$[\Lhat_{t; \alpha}]$, $\Phat_{(t)} \leftarrow \Phat_{j,k}$, $k \leftarrow k + 1$.
\ELSE
\STATE $\Phat_{(t)} \leftarrow \Phat_{(t-1)}$ 
\ENDIF
\IF{$t = \that_j + K\alpha - 1$}
\STATE $\hat{t}_{j, fin} \leftarrow t$, $\Phat_{j} \leftarrow \Phat_{(t)}$
\STATE $k \leftarrow 1$, $j \leftarrow j+1$, $\text{phase} \leftarrow \text{detect}$.
\ENDIF
\ENDIF
\IF{$\text{phase} = \text{detect}$ and $t = \hat{t}_{j-1, fin} + u\alpha$}
\STATE $\bphi \leftarrow (\I - \Phat_{j-1}\Phat_{j-1}{}')$, $\bm{B} \leftarrow \bphi\Lhat_{t, \alpha}$
\IF {$\lambda_{\max}(\bm{B}\bm{B}{}') \geq \alpha \lthres$}
\STATE $\text{phase} \leftarrow \text{update}$, $\hat{t}_j \leftarrow t$,
\ENDIF
\ENDIF 
\ENDFOR
\STATE {\bf Smoothing mode}: At $t = \that_j + K \alpha$
\textbf{for} {$t \in [\that_{j-1}+ K \alpha,  \that_j + K \alpha-1]$}  
 \\ $\Phat_{(t)}^{\mathrm{smooth}} \leftarrow \basis([\Phat_{j-1}, \Phat_{j}])$
\\  $\bpsi \leftarrow \I - \Phat_{(t)}^{\mathrm{smooth}} \Phat_{(t)}^{\mathrm{smooth}}{}'$
\\ $\lhatt^{\mathrm{smooth}} \leftarrow \yt -  \I_{\T_t} (\bpsi_{\T_t}{}'\bpsi_{\T_t})^{-1} \bpsi_{\T_t}{}' \yt$
\end{algorithmic}
\end{algorithm}
}

\Subsection{Extensions of basic NORST-miss} \label{sec:ext}




\subsubsection{Sample-Efficient-NORST-miss} This is a simple modification of NORST-miss that will reduce its sample complexity. The reason that NORST-miss needs many more observed entries is because of the projected LS step which solves for the missing entries vector, $\z_t$, after projecting $\yt$ orthogonal to $\Phat_{(t-1)}$. This step is computing the pseudo-inverse of $(\I - \Phat_{(t-1)} \Phat_{(t-1)}{}')_{\Tmisst}$. Our bound on $\missfraccol$ helps ensure that this matrix is well conditioned for any set $\Tmisst$ of size at most $\missfraccol \cdot n$.
Notice however that we prove that NORST-miss recovers $\P_j$ to $\epsilon$ accuracy with a delay of just $(K+2) \alpha = C r \log n \log(1/\epsilon)$. Once the subspace has been recovered to $\zz$ accuracy, there is no need to use projected LS to recover $\z_t$. One just needs to recover $\at$ given a nearly perfect subspace estimate and the observed entries. This can be done more easily as follows (borrows PETRELS idea): let $\Phat_{(t)} \leftarrow \Phat_{(t-1)}$, solve for $\at$ as $\hat\a_t:= (\I_{\Omega_t}{}' \Phat_{(t)})^{\dagger} \I_{\Omega_t}{}'\y_t$, and set $\lhat_t \leftarrow \Phat_{(t)} \hat\a_t$. Recall here that $\Omega_t = \Tmisst^c$. If the set of observed or missing entries was i.i.d. Bernoulli for just the later time instants, this approach will only need $\Omega (r \log r \log^2 n)$ samples at each time $t$, whp. This follows from  \cite[Lemma 3]{laura_subspace_match}.
Suppose that $\zz =1/n$, then $K \alpha = C r \log^2 n$.
Let $d_j:= t_{j+1}-t_j$ denote the duration for which the subspace is $\P_j$. Thus $\sum_j d_j = d$.  Also recall that $\rmat \le r J$.
Thus, with this approach, 
the number of observed entries needed is
$m = \Omega\left( \sum_{j=1}^J \left( n(1-c/r) K \alpha + C r \log r \log^2 n (d_j - K \alpha) \right) \right) = \Omega \left( \sum_j [ n(1-c/r) r \log^2 n + d_j r \log r \log^2 n ]\right)  = \Omega( \max(n,d) \rmat \log^2 n (\log r - c/r) )$ as long as the observed entries follow the i.i.d. Bernoulli model for the time after the first $K \alpha$ time instants after a subspace change. Or, we need the observed entries to be i.i.d. Bernoulli($1 - c/r)$ for first $K \alpha$ frames and i.i.d. Bernoulli($ r \log^2 n \log r / n$) afterwards.
Observe that the $m$ needed by sample-efficient-NORST-miss is only $(\log r - c/r)$ times larger than the best sample complexity needed by any MC technique - this is the convex methods (nuclear norm min). However sample-efficient-NORST-miss is much faster and memory-efficient compared to nuclear norm min.

\subsubsection{NORST-sliding-window} In the basic NORST approach we use a different set of estimates $\lhat_t$ for each subspace update step. So, for example, the first subspace estimate is computed at $\that_j + \alpha-1$ using $\Lhat_{\that_j+\alpha-1; \alpha}$; the second is computed at $\that_j+2 \alpha-1$ using $\Lhat_{\that_j+2\alpha-1; \alpha}$; and so on. This is done primarily to ensure mutual independence of the set of $\lt$'s in each interval because this is what makes the proof easier (allows use of matrix Bernstein for example). However, in practice, we can get faster convergence to an $\epsilon$-accurate estimate of $\P_j$, by removing this restriction. This approach is of course motivated by the sliding window idea that is ubiquitous in signal processing. For any sliding-window method, there is the window length which we keep as $\alpha$ and the hop-length which we denote by $\beta$.

Thus, NORST-sliding-window ($\beta$) is Algorithm \ref{algo:NORST-st-basic} with the following change: compute $\Phat_{j,1}$ using $\Lhat_{\that_j+\alpha-1;\alpha}$; compute $\Phat_{j,2}$ using $\Lhat_{\that_j+\alpha+\beta-1;\alpha}$; compute $\Phat_{j,3}$ using $\Lhat_{\that_j+\alpha+2\beta-1;\alpha}$;  and so on. Clearly $\beta < \alpha$ and $\beta=\alpha$ returns the basic NORST-miss.


\subsubsection{NORST-buffer} Another question if we worry only about practical performance is whether re-using the same $\alpha$ data samples $\yt$ in the following way helps: {
At $t = \that_j + k\alpha -1$, the $k$-th estimate is improved $R$ times as follows. First we obtain $\Lhat_{t;\alpha}:=[\lhat_{t-\alpha+1}, \lhat_{t-\alpha+2}, \dots \lhat_t]$ which are used to compute $\Phat_{j,k}$ via $r$-SVD. Let us denote this by $\Phat_{j,k}^{0}$. Now, we use this estimate to obtain a second, and slightly more refined estimate of the same $\L_{t;\alpha}$. We denote these as $\Lhat_{t;\alpha}^{(1)}$ and use this estimate to get $\Phat_{j,k}^{(1)}$.} This process is repeated for a total of $R + 1$ (reuse) times. We noticed that using $R=4$ suffices in most synthetic data experiments and for real data, $R=0$ (which reduces to the basic NORST algorithm) suffices.
This variant has the same memory requirement as NORST-original. The time complexity, however, increases by a factor of $R + 1$.




\section{Detailed discussion of prior art}\label{sec:prior_art}



\subsubsection{Streaming PCA with missing data, complete guarantee}
The problem of  streaming PCA with missing data was studied and a provable approach called modified block power method (MBPM) was introduced in \cite{streamingpca_miss}. 
A similar problem called ``subspace learning with partial information'' is studied in \cite{eldar_jmlr_ss}.
These give the following complete guarantee.

\begin{theorem}[streaming PCA, missing data \cite{streamingpca_miss, eldar_jmlr_ss}]
Consider a data stream, for all $t = 1, \cdots, d$, $\lt = A \zt + \wt$ where $\zt$ are $r$ length vectors generated i.i.d from a distribution $\mathcal{D}$ s.t. $\E[(\zt)_i] = 0$ and $\E[(\zt)_i^2] = 1$ and $A$ is an $n \times r$ matrix with SVD $A = \U \Lam \V{}'$ with $\lambda_1 =1 \geq \lambda_2 \geq \cdots \lambda_r = \lambda^- >0$. The noise $\wt$ is bounded: $|(\wt)_i| \leq M_{\infty}$,  and $\E[(\wt)_i^2] = \sigma^2$. Assume that (i) $A$ is $\mu$-incoherent; and (ii) we observe each entry of $\lt$ independently and uniformly at random with probability $\fracobs$; this is the Bernoulli($\rho$) model. If $\tmax \ge \alpha$ with $\alpha:=$
\begin{align*}
& {\Omega}\left(\frac{M_{\infty}^2(r \mu^2/n + \sigma^2 + nr^2(\mu^2/n + \sigma^2)^2) (\log n)^2 \log (1/\zz)  }{\log\left( \frac{\sigma^2 + 0.75 {\lambda^-}}{\sigma^2 + 0.5 {\lambda^-}}\right)(\lambda^-)^2\epsilon^2\fracobs^2} \right)
\end{align*}
then, $\SE(\Phat_{(d)}, \U) \leq \epsilon$ w.p. at least 0.99. 
\end{theorem}
There are many differences between this guarantee and ours: (i) it only recovers a single unknown subspace (since it is solving a PCA problem), and is unable to detect or track changes in the subspace; (ii) it requires the missing entries to follow the i.i.d. Bernoulli model; and (iii) it only provides a guarantee that the final subspace estimate, $\Phat_{(d)}$, is $\epsilon$-accurate (it does not say anything about the earlier estimates). (iv) Finally, even with setting $\sigma^2 = \epsilon^2 \lambda^-$ in the above (to simply compare its noise bound with ours), the required lower bound on $d$ implied by it is $d \ge C r^2 \log^2 n \log (1/\epsilon)/\fracobs^2$. This is $r \log n$ times larger than what our result requires. The lower bound on $d$ can be interpreted as the tracking delay in the setting of ST-miss. The Bernoulli model on missing entries is impractical in many settings as discussed earlier in Sec. \ref{sec:mainres_noisefree}.
On the other hand, MBPM is streaming as well as memory-optimal while our approach is not streaming and only nearly memory optimal. For a summary, see Table \ref{tab:comp_st}.
Here {\em ``streaming''} means that it needs only one pass over the data. Our approach uses SVD which requires multiple passes over short batches of data of size of order $r \log n$.


\subsubsection{ST-miss, partial guarantees}
In the ST literature, there are three well-known algorithms for {ST-miss}: PAST \cite{past,past_conv}, PETRELS \cite{petrels} and GROUSE \cite{grouse,local_conv_grouse, grouse_global,grouse_enh}. All are motivated by stochastic gradient descent (SGD) to solve the PCA problem and the Oja algorithm \cite{ojasimplified}.  These and many others are described in detail in a review article on subspace tracking \cite{chi_review}. GROUSE can be understood as an extension of Oja's algorithm on the Grassmanian. It is a very fast algorithm  since it only involves first order updates. It has been studied in \cite{grouse, local_conv_grouse, grouse_global}.
The best partial guarantee  for GROUSE rewritten in our notation is as follows. 
\begin{theorem}[GROUSE \cite{local_conv_grouse} (Theorem 2.14)]\label{thm:grouse}
Assume that the subspace is fixed, i.e., that $\P_{(t)} = \P$ for all $t$.
Denote the unknown subspace by $\P$. Let $\epsilon_t := \sum_{i=1}^r \sin^2\theta_i(\Phat_{(t)}, \P)$ where $\theta_i$ is the $i$-th largest principal angle between the two subspaces. Also, for a vector $\bm{z} \in \mathbb{R}^n$, let $\mu(\bm{z}):= \frac{n \|\bm{z}\|_{\infty}^2}{\|\bm{z}\|_{2}^2}$ quantify its denseness.
Assume that (i) $\P$ is $\mu$-incoherent; (ii) the coefficients vector $\at$ is drawn independently from a standard Gaussian distribution, i.e., $(\at)_i \overset{i.i.d.}{\sim} \mathcal{N}(0, 1)$; (iii) the size of the set of observed entries at time $t$, $\Omega_t$, satisfies $|\Omega_t| \geq  (64/3) r (\log^2 n) \mu \log(20r)$; and the following assumptions on intermediate algorithm estimates hold:
\bi
\item $\epsilon_t \leq \min(\frac{r \mu}{16n}, \frac{q^2}{128 n^2  r} )$;
\item the residual at each time, $\bm{r}_t := \lt - \Phat_{(t)}\Phat_{(t)}' \lt$ is ``dense'', i.e., it satisfies
{\small $
\mu(\bm{r}_t) \leq \min\{ \log n [\frac{0.045}{\log 10} C_1 r \mu \log(20 r)]^{0.5},
\log^2 n \frac{0.05}{8 \log 10} C_1 \log(20 r)\}
$}
with probability at least $1 - \bar{\delta}$ where $\bar{\delta} \leq 0.6$.
\ei
Then,
$
\E[\epsilon_{t+1} | \epsilon_t] \leq \epsilon_t -.32(.6 - \bar{\delta}) \frac{q}{nr}\epsilon_t + 55 \sqrt{\frac{n}{q}} \epsilon_t^{1.5}.
$
\end{theorem}
Observe that the above result makes a denseness assumption on the residual $\bm{r}_t$ and the residual is a function of $\Phat_{(t)}$. Thus it is making assumptions on intermediate algorithm estimates and hence is a partial guarantee. 

In follow-up work, the PETRELS \cite{petrels} approach was introduced. 
It is slower than GROUSE, but has much better performance in numerical experiments.
To understand the main idea of PETRELS, let us ignore the small noise $\vt$. Then, $\yt$ can be expressed as $\yt = \I_{\Omega_t} \I_{\Omega_t}{}' \lt = \I_{\Omega_t} \I_{\Omega_t}{}' \P_{(t)} \at $. Let $\tP:= \P_{(t)}$. If $\tP$ were known, one could compute $\at$ by solving a LS problem to get $\hat\a_t:= (\I_{\Omega_t}{}' \tP)^{\dagger} \I_{\Omega_t}{}'\y_t$. This of course implicitly assumes that $ \I_{\Omega_t}{}' \tP$ is well-conditioned. This matrix is of size $(n - |\Tmisst|) \times r$, thus a necessary condition for it to be well conditioned is the same as the one for NORST-miss: it also needs $n - |\Tmisst| \ge r$ although the required sufficient condition is different\footnote{If $\Omega_t$ follows an i.i.d. Bernoulli model, a sufficient condition would be $n - |\Tmisst| \ge C r \log r \log^2n$ \cite{laura_subspace_match}, or equivalently, $\missfraccol \le 1 - (Cr\log r \log^2n) /n$.}. Of course $\tP$ is actually unknown. PETRELS thus solves for $\tP$ by solving the following 
\[
\min_{\tP} \sum_{m = 1}^t \lambda^{t-m} \| \y_m -  \I_{\Omega_m} \I_{\Omega_m}{}' \tP (\I_{\Omega_m}{}' \tP)^{\dagger} \I_{\Omega_m}{}'\y_m\|^2.
\]
Here $\M^\dagger:=(\M'\M)^{-1} \M'$ and
$\lambda$ is the discount factor (set to 0.98 in their code). To solve this efficiently, PETRELS first decomposes it into updating each row of $\tilde\P$, and then parallely solves the $n$ smaller problems by second-order SGD. 
%

\begin{table*}[t!]
{
	\caption{\small{Comparing robust MC guarantees. 
We treat the condition number and incoherence parameters as constants for this table. 
}}
\begin{center}
		\resizebox{.95\linewidth}{!}{
	\begin{tabular}{@{}c@{}c@{}c@{}c@{}c@{}c@{}c}
		\toprule
\textbf{Algorithm} & \textbf{Sample complexity} & \textbf{Memory} & \textbf{Time} & \textbf{Observed entries} & \textbf{Outliers} \\
		\midrule
		NNM \cite{matcomp_first} & $\Omega(n d)$ & $\mathcal{O}(nd)$ & $\mathcal{O}(n^3/\sqrt{\epsilon})$ & i.i.d. Bernoulli ($c$) & i.i.d. Bernoulli ($c$) \\
				Projected GD \cite{rmc_gd} & $\Omega(n r^2 \log^2 n)$ & $\mathcal{O}(nd)$ & $\Omega(n r^3 \log^2 n \log^2 (1/\epsilon))$ & i.i.d. Bernoulli ($m/nd$) & bounded fraction ($\mathcal{O}(1/r)$ per row and col)  \\
NORST-miss-rob & $\Omega(nd (1 - 1/r))$ & $\mathcal{O}(n r \log n \log(1/\epsilon))$ & $\mathcal{O}(ndr \log(1/\epsilon))$ & bounded frac & bounded frac.  \\
 (this work)&  & & & $\mathcal{O}(1/r)$ per row, $\mathcal{O}(1)$ per col & $\mathcal{O}(1/r)$ per row, $\mathcal{O}(1)$ per col \\
 & & & & \multicolumn{2}{c}{Extra assumptions: Slow subspace change and lower bound on outlier magnitude} \\
	\bottomrule
	\end{tabular}
	}
	\label{tab:comp_rmc}
	\end{center}
	
	\vspace{-.2in}
	}
\end{table*}
The best guarantee for PETRELS from \cite{petrels_new} is summarized next. 
\begin{theorem}[PETRELS  \cite{petrels_new}(Theorem 2)]\label{thm:petrels}
Assume that the subspace is fixed, i.e., that $\P_{(t)} = \P$ for all $t$.
Assume that (i) the set of observed entries are drawn from the i.i.d. Bernoulli model with parameter $\rho$; (ii) the coefficients $(\at)$'s are zero-mean random vectors with diagonal covariance $\Lam$ and all higher-order moments finite; (iii) the noise, $\vt$ are i.i.d and independent of $\at$; (iv) the subspace $\P$ and the initial estimate $\Phat_0$ satisfies the following incoherence assumption
$
\sum_{i=1}^n \sum_{j=1}^r (\P)_{ij}^4 \leq \frac{C}{n},\ \text{and} \ \sum_{i=1}^n \sum_{j=1}^r (\Phat_0)_{ij}^4 \leq \frac{C}{n};
$
(v) the step-size is appropriately chosen; and (v) the initialization satisfies
$
\E\left[\|\bm{Q}_0^{(n)} - \bm{Q}(0)\|_2\right] \leq \frac{C}{\sqrt{n}}.
$
Here $\bm{Q}_0^{(n)} := \Phat_0{}' \P$ denotes the matrix of initial cosine similarities and $\bm{Q}(\tau)$ is the ``scaling limit'' which is defined as the solution of the following coupled ordinary differential equations:
\begin{align*}
\frac{d}{d\tau} \bm{Q}(\tau) = &[\fracobs \Lam^2 \bm{Q}(\tau) - 1/2 \bm{Q}(t)\bm{G}(\tau) - \\
&\bm{Q}(\tau)(\I - 1/2\bm{G}(\tau))\bm{Q}{}'(\tau)\fracobs \Lam^2\bm{Q}(\tau)]\bm{G}(\tau)\\
\frac{d}{d\tau} \bm{G}(\tau) = & \bm{G}(\tau)[ \mu - \bm{G}(\tau)(\bm{G}(\tau) + \I)(\bm{Q}{}'(\tau) \fracobs \Lam^2 \bm{Q}(\tau) + \I)]
\end{align*}
where $\fracobs$ is the subsampling ratio and $\mu = n(1-\lambda)$ where $\lambda$ is the discount parameter defined earlier.
Then, for any fixed $\tmax >0$, the time-varying cosine similarity matrix $\bm{Q}^{(n)}_{\lfloor n\tau\rfloor} = \Phat_{(\lfloor n \tau \rfloor)}{}' \P$ satisfies $\sup_{n\geq 1} \E\left[ \|\bm{Q}^{(n)}_{\lfloor n\tau\rfloor} - \bm{Q}(\tau)\| \right] \leq \frac{C_{\tmax}}{\sqrt{n}}.$
\end{theorem}
For further details, please refer to \cite[Eq's 29, 33, 34]{petrels_new}.
The above is a difficult result to further simplify since, even for $r=1$, it is not possible to obtain a closed form solution of the above differential equation. This is why it is impossible to say what this result says about $\SE(\Phat_{(t)}, \P)$  or any other error measure. Hence the above is also a {\em partial guarantee}.  \cite{petrels_new} also provides a guarantee  for GROUSE that has a similar flavor to the above result.



\subsubsection{Online MC, different model}
There are a few works with the term {\em online MC}  in their title and a reader may wrongly confuse these as being solutions to our problem.  All of them study very different ``online'' settings than ours, e.g., \cite{onlineMC1} assumes one matrix entry comes in at a time.  The work of \cite{onlineMC2} considers a problem of designing matrix sampling schemes based on current estimates of the matrix columns. This is useful only in settings where one is allowed to choose which samples to observe. This is often not possible in applications such as video analytics.

\subsubsection{MC}
%
There has been a very large amount of work on provable MC. We do not discuss everything here since MC is not the main focus of this work. 
The first guarantee for MC  was provided in \cite{matcomp_first}. This studied the nuclear norm minimization (NNM) solution. 
After NNM, there has been much later work on non-convex, and hence faster, provable solutions: alternating-minimization, e.g., \cite{optspace, lowrank_altmin, mc_luo, lowrank_altmin_no_kappa}, and projected gradient descent (proj GD), e.g., \cite{fastmc, ge_1, ge_best} and alternating-projection \cite{rmc_altproj, mc_altproj}. All these works assume a uniform random  or i.i.d. Bernoulli model on the set of missing entries (both are nearly equivalent for large $n,d$). There has been some later work that relaxes this assumption. This includes \cite{coherent_mc, noniid_mc} which assumes independent but not identical probability of the (i,j)-th entry being missed. The authors allow this probability to be inversely proportional to row and column ``leverage scores'' (quantifies denseness of a row or a column of $\L$) and hence allows the relaxing of the incoherence requirement on $\L$. If leverage scores were known, one could sample more frequently from rows or columns that are less dense (more sparse). Of course it is not clear how one could know or approximate these scores. There is also work that assumes a completely different probabilistic models on the set of observed entries, e.g., \cite{universal_mc}.
%
In summary, all existing MC works need a probabilistic model on the set of observed (equivalently, missed) entries, typically i.i.d. Bernoulli. As noted earlier this can be an impractical requirement in some applications. Our work does not make any such assumption but needs more observed entries, a detailed discussion of this is provided earlier.


\begin{algorithm}[t!]
\caption{NORST-miss-robust.
Obtain $\Phat_0$ by $C \log r$ iterations of AltProj applied to $\Y_{[1;t_\train]}$ with
$t_\train = Cr$ and with setting $(\y_t)_\Tmisst = 10$ (or any large nonzero value) for all $t=1,2,\dots,t_\train$.}
\label{algo:auto-dyn-rmc}
\begin{algorithmic}[1]
\STATE \textbf{Input}:  $\yt$, $\Tmisst$  \textbf{Output}:  $\lhatt$,  $\Phat_{(t)}$
\STATE \textbf{Extra Parameters:} $\omega_{supp} \leftarrow \smin/2 $, $\xi \leftarrow \smin/15$
\STATE $\Phat_0 \leftarrow$  obtain as given in the caption;
\STATE  $\tildej~\leftarrow~1$, $k~\leftarrow~1$, $\mathrm{phase} \leftarrow \mathrm{update}$; $\that_{0} \leftarrow t_\train$;
\FOR {$t > t_\train$}
\STATE $\bpsi \leftarrow \bm{I} - \hat{\pt}_{(t-1)}\hat{\pt}_{(t-1)}{}'$; $\tty_t \leftarrow \bpsi \yt$;
\STATE $\xhat_{t,cs} \leftarrow \arg\min_{\bm{x}} \norm{(\bm{x})_{\Tmisst^{c}}}_1 \ \text{s.t}\ \norm{\tilde{\bm{y}}_t - \bpsi \bm{x}} \leq \xi$.
\STATE $\That_t \leftarrow \Tmisst  \cup \leftarrow \{i:\ |\xhat_{t,cs}| > \omega_{supp} \}$
\STATE $\lhatt \leftarrow \yt - \I_{\That_t} ( \bpsi_{\That_t}{}' \bpsi_{\That_t} )^{-1} \bpsi_{\That_t}{}'\tty_t$
\STATE Lines $9 - 27$ of Algorithm \ref{algo:NORST-st-basic}
\ENDFOR
\STATE {\bf Offline (RMC solution): } line 25 of Algorithm \ref{algo:NORST-st-basic}. 
\end{algorithmic}
\end{algorithm}

\subsubsection{NORST for robust ST \cite{rrpcp_icml}}
While both the NORST-miss algorithm and guarantee are simple modifications of those for NORST for robust ST, our current result has two important advantages because it solves a simpler problem than robust ST.
Since there are no outliers, there is no need for the amount of subspace change or the initial estimate's accuracy to be smaller than the outlier magnitude lower bound.  This was needed in the robust ST case to obtain an estimate of the outlier support $\Tt$. Here, this support is known.
This is why NORST-miss has the following two advantages.
(i) It works with a zero initialization where as NORST (for robust ST) required a good enough initialization for which AltProj or PCP needed to be applied on an initial short batch of observed data. (ii) It does not need an upper bound on the amount of subspace change at each $t_j$, it allows both slow and sudden changes.



\begin{table*}[ht!]
{
	\caption{\small{(top) Number of samples (frames) required by NORST and its heuristic extensions, and PETRELS to attain $\approx 10^{-16}$ accuracy. The observed entries are drawn from a i.i.d. Bernoulli model with $\rho = 0.7$ fraction of observed entries. Notice that NORST-buffer($4$) and NORST-sliding-window ($\beta=10, R=1$) converges at the same rate as PETRELS and the time is also comparable. The other variants require more samples to obtain the same error but are faster compared to PETRELS. (bottom) Evaluation of Sample Efficient NORST with $\rho_1 = 0.9$ and $\rho_2 = 0.15$.}}
	\begin{center}
			\begin{tabular}{  c c c c c c c c c }
				\toprule
				Algorithm & NORST & \multicolumn{4}{c}{NORST-buffer} & \multicolumn{2}{c}{NORST-sliding-window and buffer} & PETRELS \\ \cmidrule(lr){1-1} \cmidrule(lr){2-2}  \cmidrule(lr){3-6} \cmidrule(lr){7-8} \cmidrule(lr){9-9}
	Parameter $R$, $\beta$		&		&  $R=1$ & $R=2$ & $R=3$ & $R=4$ & $\beta = 1$, $R = 0$ & $\beta = 10$, $R=1$ &  \\
	Time taken (ms)			 & $1.9$ & $10.8$ & $18.6$ & $27.5$ & $34.5$ & $16$ & $35$ & $33$ \\
				Number of samples & $3540$ & $2580$ & $2100$ & $2050$ & $1950$ & $2400$ & $1740$ & $1740$ \\
				\bottomrule
			\end{tabular}
	\end{center}
	\label{tab:convergence}
\begin{center}
					\resizebox{.8\linewidth}{!}{
			\begin{tabular}{c c c c c}
				\toprule
				Algorithm & NORST-miss ($6$)& NORST-samp-eff ($ 1$) & PETRELS ($15$) & GROUSE ($2$)\\  \midrule
				Average Error & ${0.04}$  & ${0.04}$  & $0.02$  & $0.13$ \\
				\bottomrule
			\end{tabular}
		}

\end{center}	
	}
	\vspace{-.2in}
\end{table*}

\section{Robust ST with missing entries} \label{sec:norstmissrob}

Robust ST with missing entries (RST-miss) is a generalization of robust ST and of ST-miss. 
In this case, we observe $n$-dimensional data vectors that satisfy
\begin{align}
\yt = \proj(\lt + \gt) + \vt, \text{ for } t = 1, 2, \dots, \tmax.
\label{eq:rmc_prob}
\end{align}
where $\gt$'s are the sparse outliers.  
Let $\xt := \proj(\gt)$. We use $\Tspart$ to denote the support of $\xt$. This is the part of the outliers that actually corrupt our measurements, thus in the sequel we will only work with $\xt$.
With $\xt$ defined as above, $\yt$ can be expressed as
\begin{align}
\yt = \proj(\lt) + \xt + \vt
\end{align}
Observe that, by definition, $\xt$ is supported outside of $\Tmisst$ and hence $\Tmisst$ and $\Tspart$ are disjoint.
%
Defining the $n \times \tmax$ matrix $\L:= [\l_1, \l_2, \dots \l_d]$, the above is a robust MC problem.

The main modification needed in this case is outlier support recovery. The original NORST for robust ST  \cite{rrpcp_icml} used $l_1$ minimization followed by thresholding based support recovery for this purpose. In this case, the combined sparse vector is $\tilde{\x}_t:= \xt - \I_\Tmisst \I_\Tmisst{}' \lt$. Support recovery in this case is thus a problem of sparse recovery with partial support knowledge $\Tmisst$.  In this case, we can still use $l_1$ minimization followed by thresholding. However a better approach is to use noisy modified-CS \cite{modcsjournal,stab_jinchun_jp} which was introduced to exactly solve this problem. We use the latter. 
The second modification needed is that, just like in case of robust ST, we need an accurate subspace initialization. To get this, we can use the approach used in robust ST \cite{rrpcp_icml}: for the initial $C r \log n \log (1/\zz)$ samples, use the AltProj algorithm for robust PCA (while ignoring the knowledge of $\Tmisst$ for this initial period). We summarize the approach in Algorithm \ref{algo:auto-dyn-rmc}.


We have the following guarantee for NORST-miss-robust.  Let $\outfracrow_{\alpha}$ be the maximum fraction of outliers per row of any sub-matrix of $\X$ with $\alpha$ consecutive columns; $\outfraccol$ be the maximum fraction of outlier per column of $\X$.
Also define $\xmint:=\min_t \min_{i \in \Tspart} |(\xt)_i|$ to denote the minimum outlier magnitude and let $\dif:=\max_j \Delta_j = \max_j \SE(\P_{j-1}, \P_j)$.
\begin{corollary}
Consider Algorithm \ref{algo:auto-dyn-rmc}. Assume all conditions of Theorem \ref{thm:stmiss} hold and
\ben
%

\item $\missfraccol + 2\cdot \outfraccol \le  \frac{c_1}{\mu r}$; and $\missfracrow_{\alpha} + \outfracrow_{\alpha} \le  \frac{c_2}{f^2}$;
\item subspace change:
\ben
\item $t_{j+1}-t_j > (K+2)\alpha$, and
\item $\dif \le 0.8$ and $C_1 \sqrt{ r \lambda^+} (\Delta + 2 \zz) \leq \xmint$
\een

\item initialization satisfies $\SE(\Phat_0,\P_0) \le 0.25$ and $C_1 \sqrt{r \lambda^+}  \SE(\Phat_0,\P_0) \le \xmint$;%
\een
then, all guarantees of Theorem \ref{thm:stmiss} and Corollary \ref{cor:thm1} hold.
\label{cor:dyn_rmc}
\end{corollary}

{
\begin{remark}[Relaxing outlier magnitudes lower bound]
As also explained in \cite{rrpcp_icml}, the outlier magnitude lower bound can be significantly relaxed.
First, without any changes, if we look at the proof, our required lower bound on outlier magnitudes is actually $0.3^{k-1} \sqrt{ r \lambda^+} (\Delta + 2 \zz)$ in interval $k$ of subspace update. To be precise, we only need $\min_{t \in \J_k} \min_{i \in \Tspart} |(\xt)_i| \ge 0.3^{k-1} \sqrt{ r \lambda^+} (\Delta + 2 \zz)$. Here $\J_k$ is the interval defined in Theorem \ref{thm:stmiss}. Thus, for $t \in \J_{K+1}$ (after the update step is complete but the subspace has not changed), we only need $ \min_{i \in \Tspart} |(\xt)_i| \ge \zz \sqrt{ r \lambda^+}$.
Moreover, this can be relaxed even more as explained in Remark 2.4 of \cite{rrpcp_icml}.
\end{remark}
}
The proof is similar to that given in \cite{rrpcp_icml}. Please see the Appendix for an explanation of the differences. The advantage of using modified-CS to replace $l_1$ min when recovering the outlier support is that it weakens the required upper bound on $\missfraccol$ by a factor of two. If we used $l_1$ min, we would need $2 \cdot (\missfraccol + \outfraccol)$ to satisfy the upper bound given in the first condition. 

\subsubsection{Comparison with existing work}
Existing solutions for robust ST-miss include GRASTA \cite{grass_undersampled}, APSM \cite{chouvardas2015robust} and ROSETA \cite{mansour_robust_ss_track}. APSM comes with a partial guarantee, while GRASTA and ROSETA do not have a guarantee.
The first few provable guarantees for robust MC were \cite{rpca, ranksparSanghavi}. Both studied the convex optimization solution
which was  slow. 
Recently, there have been two other works \cite{rpca_gd, rmc_gd}  which are projected-GD based approaches and hence are much faster. These assume an $\mathcal{O}(1/r)$ bound on outlier fractions per row and per column. All these assume that the set of observed entries is i.i.d. Bernoulli. 

Compared with these, our result needs slow subspace change and a lower bound on outlier magnitudes; but it does not need a probabilistic model on the set of missing or outlier entries, and improves the required upper bound on outlier fractions per row by a factor of $r$. Also, our result needs more observed entries in the setting of $\rmat \approx r$, but not when $\rmat$ is significantly larger than $r$, for example not when $\rmat$ is nearly linear in $d$. A summary of this discussion is given in Table \ref{tab:comp_rmc}.




%

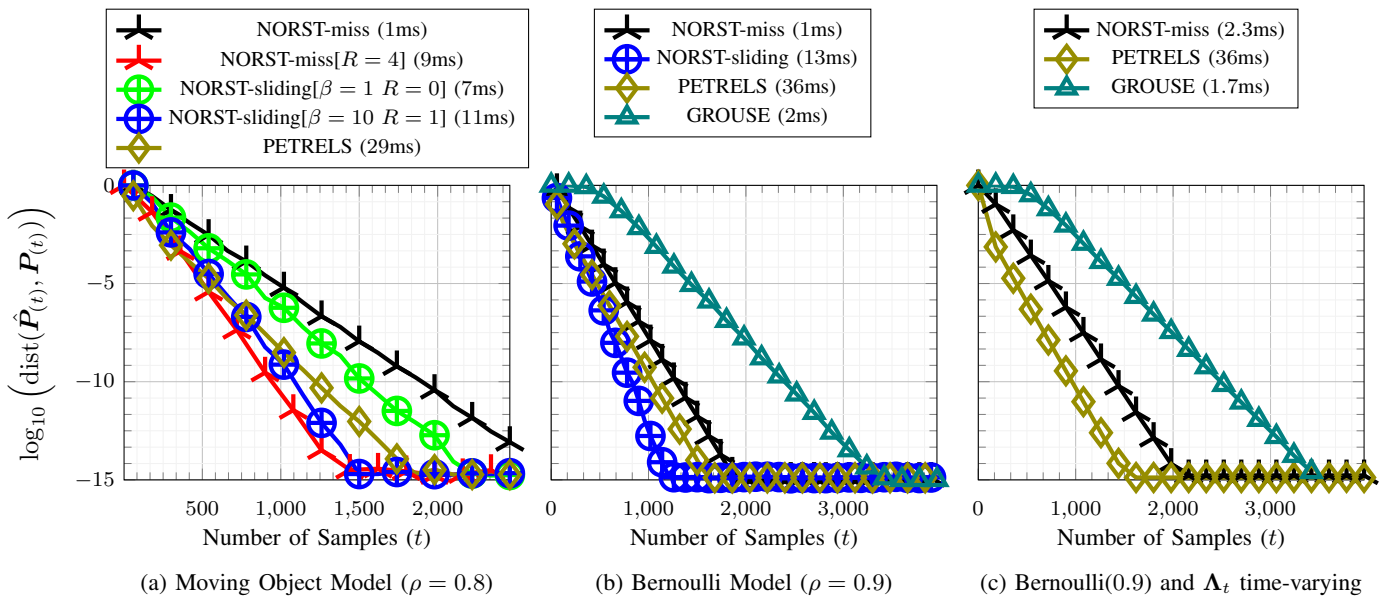
\begin{figure*}[t!]
\centering
\begin{tikzpicture}
    \begin{groupplot}[
        group style={
            group size=3 by 1,
            horizontal sep=.55cm,
            vertical sep=1cm,
                        x descriptions at=edge bottom,
            y descriptions at=edge left,
        },
                ymin=-15, ymax=0,
        enlargelimits=false,
        width = .37\linewidth,
        height=5.5cm,
        enlargelimits=false,
                grid=both,
    grid style={line width=.1pt, draw=gray!10},
    major grid style={line width=.2pt,draw=gray!50},
    minor x tick num=5,
    minor y tick num=5,
    ]
       \nextgroupplot[
		legend entries={
					NORST-miss ($1$ms),
					NORST-miss[$R=4$] ($9$ms),
            		NORST-sliding[$\beta=1$ $R=0$] ($7$ms),
            		NORST-sliding[$\beta=10$ $R=1$] ($11$ms),            		
            		PETRELS ($29$ms),
            		},
            legend style={at={(1.05, 1.6)}},
            legend columns = 1,
            legend style={font=\footnotesize},
            xlabel={\small{Number of Samples ($t$)}},
            ylabel={{$\log_{10}\left(\SE(\Phat_{(t)},\P_{(t)})\right)$}},
                        title style={at={(0.5,-.35)},anchor=north,yshift=1},
            title={\small{(a) Moving Object Model ($\rho = 0.8$)}},
                        xticklabel style= {font=\footnotesize, yshift=-1ex},
            yticklabel style= {font=\footnotesize},
        ]
                	        \addplot [black, line width=1.6pt, mark=Mercedes star,mark size=6pt, mark repeat=2] table[x index = {0}, y index = {1}]{\mofix};
	        \addplot [red, line width=1.6pt, mark=Mercedes star,mark size=6pt, select coords between index={0}{13}] table[x index = {2}, y index = {3}]{\mofix};
	        \addplot [green, line width=1.6pt, mark=oplus,mark size=5pt, mark repeat=2] table[x index = {4}, y index = {5}]{\mofix};
	        \addplot [blue, line width=1.6pt, mark=oplus,mark size=5pt, mark repeat=2] table[x index = {6}, y index = {7}]{\mofix};
			\addplot [olive, line width=1.6pt, mark=diamond,mark size=5pt, mark repeat=2, select coords between index={0}{23}] table[x index = {8}, y index = {9}]{\mofix};
	               \nextgroupplot[
	               		legend entries={
					NORST-miss ($1$ms),
            		NORST-sliding ($13$ms),
            		PETRELS ($36$ms),
            		GROUSE ($2$ms)
            		},
            legend style={at={(.83,1.6)}},
            legend columns = 1,
            legend style={font=\footnotesize},
            xlabel={\small{Number of Samples ($t$)}},
                        title style={at={(0.5,-.35)},anchor=north,yshift=1},
            title={\small{(b) Bernoulli Model ($\rho = 0.9$)}},
                        xticklabel style= {font=\footnotesize, yshift=-1ex},
                                    yticklabel style= {font=\footnotesize},
        ]
                	        \addplot [black, line width=1.6pt, mark=Mercedes star,mark size=6pt, mark repeat=2] table[x index = {0}, y index = {1}]{\bernfix};
	        \addplot [blue, line width=1.6pt, mark=oplus,mark size=5pt, mark repeat=2, select coords between index={0}{64}] table[x index = {2}, y index = {3}]{\bernfix};
	        \addplot [olive, line width=1.6pt, mark=diamond,mark size=5pt, mark repeat=3] table[x index = {4}, y index = {5}]{\bernfix};
	        \addplot [teal, line width=1.6pt, mark=triangle,mark size=4pt, mark repeat=1, select coords between index={0}{22}] table[x index = {6}, y index = {7}]{\bernfix};
	
	        	               \nextgroupplot[
	        	               		legend entries={
					NORST-miss ($2.3$ms),
            		PETRELS ($36$ms),
            		GROUSE ($1.7$ms),
            		},
            legend style={at={(.83,1.6)}},
            legend columns = 1,
            legend style={font=\footnotesize},
            xlabel={\small{Number of Samples ($t$)}},
                                                title style={at={(0.5,-.35)},anchor=north,yshift=1},
            title={\small{(c) Bernoulli($0.9$) and $\Lam_t$ time-varying}},
                        xticklabel style= {font=\footnotesize, yshift=-1ex},
            yticklabel style= {font=\footnotesize},
        ]
                	        \addplot [black, line width=1.6pt, mark=Mercedes star,mark size=6pt, mark repeat=1] table[x index = {0}, y index = {1}]{\changelambda};
	        \addplot [olive, line width=1.6pt, mark=diamond,mark size=5pt, mark repeat=1] table[x index = {2}, y index = {3}]{\changelambda};
	        \addplot [teal, line width=1.6pt, mark=triangle,mark size=4pt, mark repeat=1, select coords between index={0}{19}] table[x index = {4}, y index = {5}]{\changelambda};
	
    \end{groupplot}
\end{tikzpicture}
\caption{\small{We compare NORST-miss and its extensions with PETRELS and GROUSE. We plot the logarithm of the subspace error between the true subspace $\P_{(t)}$ and the algorithm estimates, $\Phat_{(t)}$ on the y-axis and the number of samples ($t$) on the x-axis. As can be seen, in the first two cases, NORST-buffer and NORST-sliding have the best performance (while also being faster than PETRELS), followed by PETRELS, basic NORST and then GROUSE. PETRELS performs best in the scenario of time varying $\Lam_t$. The computational time per sample (in milliseconds) for each algorithm is mentioned in the legend.
}}
\label{fig:fixed_ss}
\end{figure*}

\section{Experimental Comparisons} \label{sec:sims_main}

We present the results of numerical experiments on synthetic and real data\footnote{We downloaded the PETRELS' and GROUSE code from the authors' website and all other algorithms from \url{https://github.com/andrewssobral/lrslibrary}.}. All the codes for our experiments are available at \url{https://github.com/vdaneshpajooh/NORST-rmc}.  {\em In this section, we refer to NORST-miss as just NORST.} All time comparisons are performed on a Desktop Computer with Intel Xeon E3-1200 CPU, and 8GB RAM.

{
\Subsection{Parameter Setting for NORST} \label{sec:sims_param}
The algorithm parameters required for NORST are $r$, $K$, $\alpha$ and $\lthres$. For our theory, we assume $r$, $\lambda^+$, $\lambda^-$, are known, and we pick a desired accuracy, $\epsilon$. We set $K = C \log(1/\epsilon)$, $\alpha = Cf^2 r \log n$, and $\lthres = 2 \epsilon^2 \lambda^-$ with $C$ being a numerical constant more than one.
Experimentally, the value of $r$ needs to be set from model knowledge, however, overestimating it by a little does not significantly affect the results. In most of our experiments, we set $\alpha = 2r$ (ideally it should grow as $r \log n$ but since $\log n$ is very small for practical values of $n$ it can be ignored). 
 $\alpha$ should be a larger multiple of $r$ when either the data is quite noisy or when few entries are observed.
We set $K$ based on how accurately we would like to estimate the subspace.
%
The parameter $\lthres$ needs to be set as a small fraction of the minimum signal space eigenvalue. In all synthetic data experiments, we set $\lthres = 0.0008 \lambda^-$. Another way to set $\lthres$ is as follows. After $K\alpha$ frames, we can estimate $\hat{\lambda}^-$ as the $r$-th eigenvalue of $\sum_{\tau = t-\alpha+1}^t \lhat_\tau \lhat_\tau{}' / \alpha$ and set $\lthres = c \hat{\lambda}^-$ as mentioned before. We use the Conjugate Gradient Least Squares (CGLS) method \cite{cgls} for the LS step with tolerance as $10^{-16}$, and maximum iterations as $20$.

For the video experiments, we estimated $r$ using training data from a few videos and fixed it as $r=30$. We let  $\lambda^-$ be the $r$-th eigenvalue of the training dataset.
We used $\lthres = 1.6 \times 10^{-6} \lambda^- = 0.002$, $\alpha = 2r$ and $K = 3$ for the video data.
The reason that we use a smaller fraction of $\lambda^-$ as $\lthres$ is because videos are only approximately low-rank. 
}

\Subsection{Fixed Subspace, Noise-free data} \label{sec:sims_fixed}
We generated the data according to \eqref{orpca_eq} and set $\vt = 0$. We assume a fixed subspace i.e. $J=1$. We generate the subspace basis matrix $\P \in \mathbb{R}^{n \times r}$ by ortho-normalizing the columns of a random Gaussian matrix with $n = 1000$ and $r = 30$. The $\at$'s (for $t = 1, \cdots, d$ and $d = 4000$) are generated independently as $(\a_t)_i \stackrel{\text{i.i.d}}{\sim} \text{unif}[-q_i, q_i]$ where $q_i = \sqrt{f} - \sqrt{f}(i-1)/2r \quad \text{for} \quad i = 1, 2, \cdots, r-1$ and $q_r = 1$. Thus, the condition number of $\Lam$ is $f$ and we set $f = 100$.

For our first experiment,  the observed entries' set was i.i.d. Bernoulli with fraction of observed entries  $\rho=0.7$. We compared all NORST extensions and PETRELS. We set the algorithm parameters for NORST and extensions as mentioned before and used $K = 33$ to see how low the NORST error can go. For PETRELS we set \texttt{max$\_$cycles} $=1$, forgetting parameter $\lambda = 0.98$ as specified in the paper. We display the results in Table \ref{tab:convergence} (top). Notice that NORST-miss and its extensions are significantly faster than PETRELS. Also, the $\beta=10,R=1$ is the best of all the NORST extensions and is as good as PETRELS.

{

\begin{figure*}[t!]
\centering
\begin{tikzpicture}
    \begin{groupplot}[
        group style={
            group size=3 by 1,
            horizontal sep=.4cm,
            vertical sep=1cm,
                        x descriptions at=edge bottom,
           y descriptions at=edge left,
        },
        enlargelimits=false,
        width = .37\linewidth,
        height=6cm,
        ymin=1e-10, ymax=1e0,
                grid=both,
    grid style={line width=.1pt, draw=gray!10},
    major grid style={line width=.2pt,draw=gray!50},
    minor tick num=5,
    ]
       \nextgroupplot[
		legend entries={
					NORST-miss ($3.1$ms),
            		NORST-sliding ($5.8$ms),            		
            		PETRELS ($35$ms),
            		GROUSE ($2.9$ms)
            		},
            legend style={at={(2.8, 1.2)}},
            legend columns = 4,
            legend style={font=\footnotesize},
            ymode=log,
            xlabel={\small{Number of Samples ($t$)}},
            ylabel={{$\SE(\Phat_{(t)},\P_{(t)})$}},
            title style={at={(0.5,-.3)},anchor=north,yshift=1},
            title={\small{(a) Piecewise Constant (Noisy)}},
                        xticklabel style= {font=\footnotesize, yshift=-1ex},
            yticklabel style= {font=\footnotesize},
        ]
                	        \addplot [black, line width=1.2pt, mark=Mercedes star,mark size=6pt, mark repeat=1, select coords between index={0}{15}] table[x index = {0}, y index = {1}]{\changenoise};
	        \addplot [red, line width=1.2pt, mark=square,mark size=3pt, mark repeat=1] table[x index = {2}, y index = {3}, select coords between index={0}{15}]{\changenoise};
	        \addplot [olive, line width=1.2pt, mark=o,mark size=4pt, mark repeat=1] table[x index = {4}, y index = {5}, select coords between index={0}{15}]{\changenoise};
	        \addplot [teal, line width=1.2pt, mark=diamond,mark size=5pt, mark repeat=1] table[x index = {6}, y index = {7}, select coords between index={0}{15}]{\changenoise};
	
	               \nextgroupplot[
            legend style={at={(1.2, 1.6)}},
            legend columns = 3,
            legend style={font=\footnotesize},
            ymode=log,
            xlabel={\small{Number of Samples ($t$)}},
                        xticklabel style= {font=\footnotesize, yshift=-1ex},
            yticklabel style= {font=\footnotesize},
                        title style={at={(0.5,-.3)},anchor=north,yshift=1},
                        title={\small{(b) Piecewise Constant (Noise-Free)}},
        ]
			\addplot [black, line width=1.2pt, mark=Mercedes star,mark size=6pt, mark repeat=1] table[x index = {0}, y index = {1}, select coords between index={0}{24}]{\pwjustall};

			\addplot [olive, line width=1.2pt, mark=o,mark size=4pt, mark repeat=1] table[x index = {6}, y index = {7}, select coords between index={0}{24}]{\pwjustall};

			\addplot [teal, line width=1.2pt, mark=diamond,mark size=5pt, mark repeat=1] table[x index = {12}, y index = {13}, select coords between index={0}{24}]{\pwjustall};


	               \nextgroupplot[
            legend style={at={(.8, 1.6)}},
            legend columns = 1,
            legend style={font=\footnotesize},
            ymode=log,
            xlabel={\small{Number of Samples ($t$)}},
                        title style={at={(0.5,-.3)},anchor=north,yshift=1},
            title={\small{(c) Subspace change at each time}},
                        xticklabel style= {font=\footnotesize, yshift=-1ex},
            yticklabel style= {font=\footnotesize},
        ]
                	        \addplot [black, line width=1.2pt, mark=Mercedes star,mark size=6pt, mark repeat=2] table[x index = {0}, y index = {1}]{\changetime};
	        \addplot [olive, line width=1.2pt, mark=o,mark size=4pt, mark repeat=1] table[x index = {2}, y index = {3}]{\changetime};
	        \addplot [teal, line width=1.2pt, mark=diamond,mark size=5pt, mark repeat=1] table[x index = {4}, y index = {5}]{\changetime};
    \end{groupplot}
\end{tikzpicture}
\caption{\small{Subspace error versus time plot for changing subspaces. We plot the $\SE(\Phat_{(t)},\P_{(t)})$ on the y-axis and the number of samples ($t$) on the x-axis.  The entries are observed under Bernoulli model with $\rho=0.9$. The computational time taken per sample (in milliseconds) is provided in the legend parenthesis. {\bf (a)  Piecewise constant subspace change and noise-sensitivity:} 
Observe that after the first subspace change, NORST-sliding adapts to subspace change using the least number of samples and is also $\approx$ 6x faster than PETRELS whereas GROUSE requires more samples than our approach and thus is unable to converge to the noise-level ($\approx 10^{-4}$); {\bf (b) Piecewise Constant and noise-free:} All algorithms perform significantly better since the data is noise-free. We clip the y-axis at $10^{-10}$ for the sake of presentation but NORST and PETRELS attain a recovery error of $10^{-14}$. {\bf (c) Subspace changes a little at each time:} All algorithms are able to track the span of top-$r$ singular vectors of $[\P_{(t-\alpha+1)}, \cdots , \P_{(t)}]$ to an accuracy of $10^{-4}$. As explained, the subspace change at each time can be thought of as noise. GROUSE needs almost $2$x number of  samples to obtain the same accuracy as NORST while PETRELS is approximately $10$x slower than both NORST and GROUSE. }}
\label{fig:stmiss}
\end{figure*}
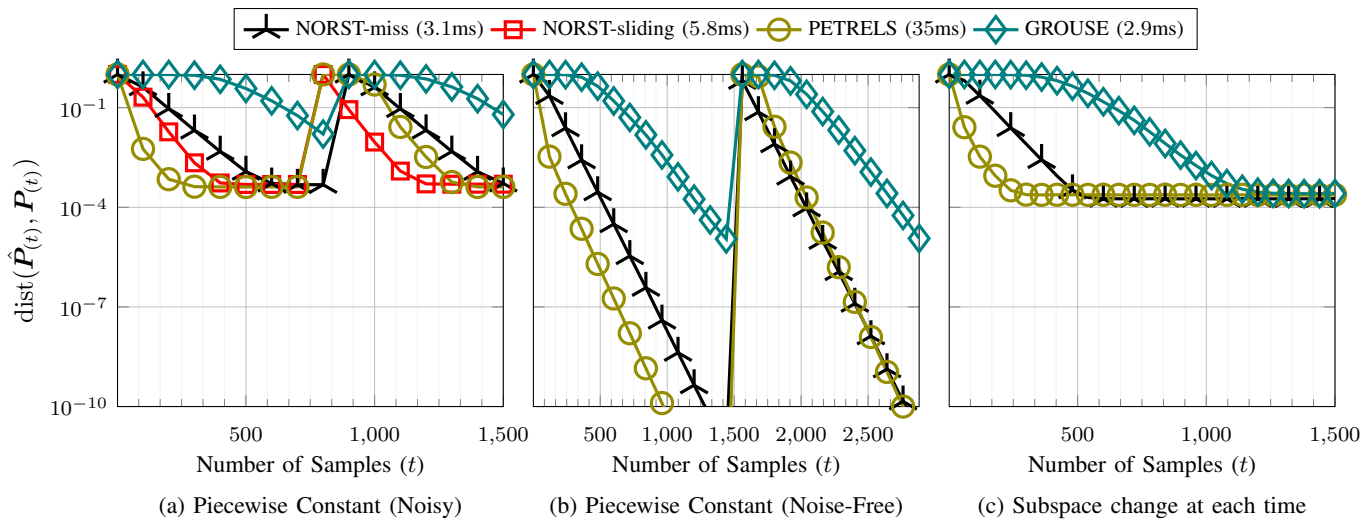


In our second set of experiments, we compared NORST (and a few extensions) with PETRELS and GROUSE for three settings of missing data.
For GROUSE, we set maximum cycles as $1$ as specified in the documentation and set the step size, $\eta = 0.1$ and the step-size is udpated according to \cite{grouse_global}.
The first was for missing generated from the  Moving Object model \cite[Model 6.19]{rrpcp_dynrpca} with $s = 200$, and $b_0 = 0.05$. This translates to $\rho = 0.8$ fraction of observed entries. This is an example of a deterministic model on missing entries. We plot the subspace recovery error versus time for this case in Fig. \ref{fig:fixed_ss}(a) As can be seen, NORST-buffer (R=4) and NORST-sliding-window ($\beta=10,R=4$) have the best performance, followed by PETRELS, basic NORST, and then GROUSE. PETRELS is the slowest in terms of time taken. In Fig.  \ref{fig:fixed_ss}(b), we  plot the results for Bernoulli observed entries' set with $\rho=0.9$. Here again, NORST-sliding has the best performance. Basic NORST is only slightly worse than PETRELS.
{
 As can be seen from the time taken (displayed in the legend), NORST and its extensions are much faster than PETRELS.}

In Fig. \ref{fig:fixed_ss}(c), as suggested by an anonymous reviewer, we evaluate the same case but with the covariance matrix of $\lt$ being time-varying. We generate the $\at$'s as described earlier but with  $q_{t,i} = \sqrt{f} - \sqrt{f}(i-1)/2r - \lambda^-/2$ for $t = 2, 4, 6,\cdots$ and $q_{t,i} = \sqrt{f} - \sqrt{f}(i-1)/2r + \lambda^-/2$ for $t = 1, 3, 5,\cdots $ and $q_{t,r} = 1$. As can be seen all approaches still work in this case. PETRELS converges with the fewest samples but is almost $18x$ slower.

\Subsection{Changing Subspaces, Noisy and Noise-free Measurements} \label{sec:sims_change}

{
\textbf{{Piecewise constant subspace change, noisy and noise-free:}} We generate the changing subspaces using $\P_j = e^{\gamma_j \B_j} \P_{j-1}$ as done in \cite{chi_review} where $\gamma_j$ controls the amount subspace change and $\B_j$'s are skew-symmetric matrices. We used the following parameters: $n = 1000$, $d = 10000$, $J = 6$, and the subspace changes after every $800$ frames. The other parameters are $r = 30$, $\gamma_j =  100$ and the matrices $\B_i$ are generated as $\B_i = (\tilde{\B}_i - \tilde{\B}_i{}')$ where the entries of $\tilde{\B}_i$ are generated independently from a standard normal distribution and $\at$'s are generated as in the fixed subspace case. For the missing entries supports, we consider the Bernoulli Model with $\rho = 0.9$.
The noise $\vt$'s are generated as i.i.d. Gaussian r.v.'s with $\sqrt{\lambda_v^+}= 3 \times 10^{-3} \sqrt{\lambda^-}$. The results are summarized in Fig. \ref{fig:stmiss}(a). 
For NORST we set $\alpha = 100$ and $K=7$. We observe that all algorithms except GROUSE are able to attain final accuracy approximately equal to the noise-level, $10^{-3}$ within a short delay of the subspace change. 
We also observe that NORST-sliding-window adapts to subspace change using the fewest samples possible. Moreoever it is much faster than PETRELS.
}

In Fig. \ref{fig:stmiss}(b), we plot results for the above setting but with noise $\nu_t=0$. In this case, the underlying subspace is recovered to accuracy lower than $10^{-12}$ by NORST and PETRELS but GROUSE only tracks to error $10^{-7}$.

\begin{table*}[t!]
	{
		\caption{\small{Comparison of $\|\L - \hat{\L}\|_F/ \|\L\|_F$ for MC. We report the time taken per sample in milliseconds in parenthesis. Thus the table format is Error (computational time per sample).
The first three rows are for the fixed subspace model. The fourth row contains results for time-varying subspace and with noise of standard deviation $0.003 \sqrt{\lambda^-}$ added. The last row reports Background Video Recovery results (for the curtain video shown in Fig. \ref{fig:vid_mo_st} when missing entries are Bernoulli with $\rho=0.9$.}}
		\begin{center}
			\resizebox{.95\linewidth}{!}{
				\begin{tabular}{  c c c c c} \toprule
						Subspace model & {NORST-smoothing} & \multicolumn{2}{c}{nuclear norm min (NNM) solvers} & projected-GD  \\ \cmidrule(lr){3-4}
						& & IALM & SVT &  \\ \midrule
Fixed (Bern, $\rho=0.9$) &\textbf{$1.26 \times 10^{-15}$ ($10$)} & $1.43 \times 10^{-12}$ ($150$) & $7.32 \times 10^{-7}$ ($164$) & $0.98$ ($1$) \\		
Fixed (Bern, $\rho=0.3$) &	$3.5 \times 10^{-6}$ ($11$) & $5.89 \times 10^{-13}$ ($72$) &--   & $0.98$ ($9$) \\
Noisy, Changing (Bern, $\rho=0.9$) & $3.1 \times 10^{-4}$ ($3.5$) & $3.47 \times 10^{-4}$ ($717$) & $2.7 \times 10^{-3}$ ($256$) &   $0.97$ ($2$)  \\			
Video Data	& $0.0074$ ($83.7$) &	$0.0891$ ($57.5$) &	$0.0034$ ($6177 $) & --	\\

	\bottomrule
				\end{tabular}
			}
		\end{center}
		\label{tab:all_MCalgos_frob}
	}
	\vspace{-.2in}
\end{table*}

{
\textbf{Subspace change at each time:} 
Here we generate the data using the approach of \cite{grouse}: $\P_{(1)}$ is generated by ortho-normalizing the columns of a i.i.d. Gaussian matrix and let $\P_{(t)} = e^{\gamma \B} \P_{(t-1)}$. We set $\gamma = 10^{-7}$. No extra noise $\vt$ was added, i.e., $\vt=0$, in this experiment. We plot $\SE(\Phat_{(t)}, \P_{(t)})$ in Fig. \ref{fig:stmiss}(c).
Notice that, even without added noise $\vt$, all algorithms are only able to track the subspaces to accuracy at most $10^{-3}$ in this case. 
The reason is, as explained earlier in Sec. \ref{identif}, subspace change at each time can be interpreted as $r$ dimensional piecewise constant subspace change plus noise.
}

\begin{figure*}[t!]
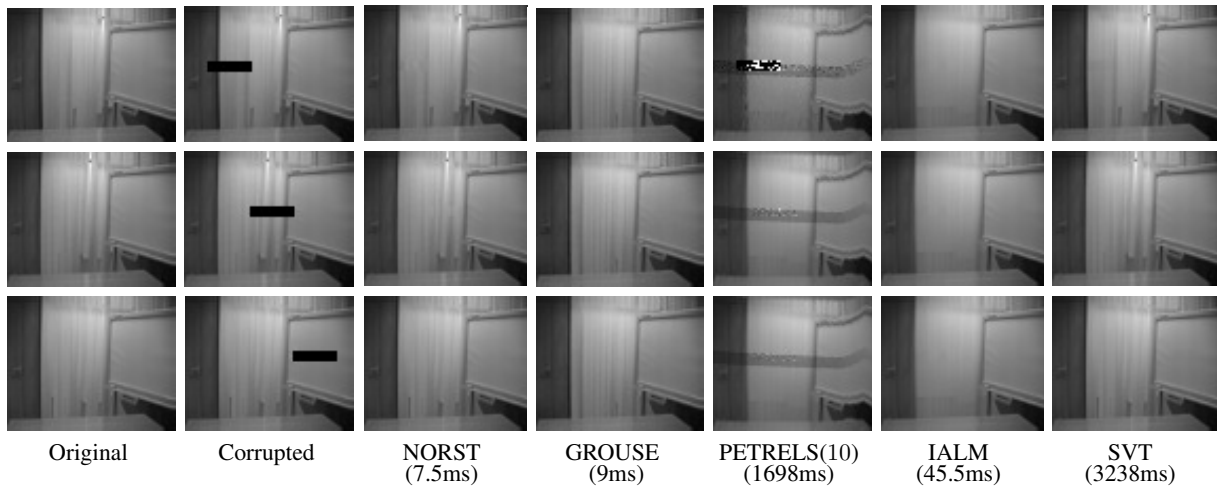

\centering
\resizebox{.9\linewidth}{!}{
\begin{tabular}{@{}c @{}c @{}c @{}c @{}c @{}c @{}c @{}}
\movobjorigFrames & \movobjcorruptedFrames & \movobjnorstFrames & \movobjgrouseFrames & \movobjpetrelsXFrames & \movobjialmFrames & \movobjsvtFrames \\
\scalebox{1.5}{Original}  & \scalebox{1.5}{Corrupted}  &\scalebox{1.5}{NORST}  & \scalebox{1.5}{GROUSE } & \scalebox{1.5}{PETRELS($10$)}  & \scalebox{1.5}{IALM}  & \scalebox{1.5}{SVT} \\
& &\scalebox{1.5}{(7.5ms)}  & \scalebox{1.5}{(9ms)} & \scalebox{1.5}{(1698ms)}  & \scalebox{1.5}{(45.5ms)}  & \scalebox{1.5}{(3238ms)}
\end{tabular}
}
\caption{\small{Background Recovery under Moving Object Model missing entries ($\rho = 0.98$). We show the original, observed, and recovered frames at $t = \{980, 1000, 1020\}$. NORST and SVT are the only algorithms that work although NORST is almost $3$ orders of magnitude faster than SVT. PETRELS($10$) exhibits artifacts, while IALM and GROUSE do not capture the movements in the curtain. The time taken per sample for each algorithm is shown in parenthesis.}}
\label{fig:vid_mo_st}
\end{figure*}

\begin{figure*}[t!]
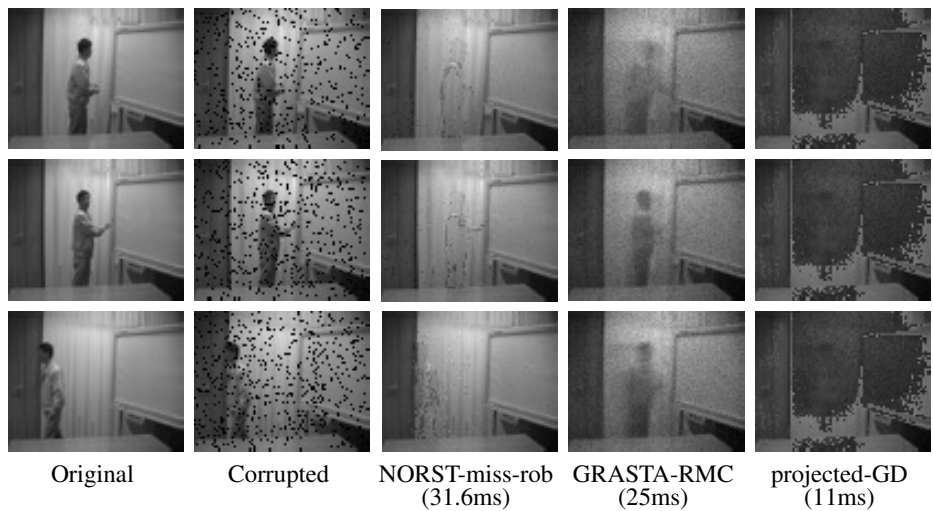

\centering
\resizebox{.7\linewidth}{!}{
\begin{tabular}{@{}c @{}c @{}c @{}c @{}c}
\bgfgorigFrames & \bgfgcorruptedFrames & \bgfgNORSTFrames &\bgfgGRASTAFrames & \bgfgNCRMCFrames \\
\scalebox{1.5}{Original}  & \scalebox{1.5}{Corrupted}  &\scalebox{1.5}{NORST-miss-rob}  & \scalebox{1.5}{GRASTA-RMC}  & \scalebox{1.5}{projected-GD} \\
&  &\scalebox{1.5}{(31.6ms)}  & \scalebox{1.5}{(25ms)}  & \scalebox{1.5}{(11ms)}

\end{tabular}
}
\caption{\small{Background Recovery with foreground layer, and Bernoulli missing entries ($\rho = 0.9$). We show the original, observed and recovered frames at $t = 1755+\{1059, 1078, 1157\}$. NORST-miss-rob exhibits artifacts, but is able to capture most of the background information, whereas, GRASTA-RMC and projected-GD fail to obtain meaningful estimates. The time taken per sample for each algorithm is shown in parenthesis.}}
\label{fig:vid_rmc}
\end{figure*}

\Subsection{Matrix Completion} \label{sec:sims_mc}
{
In Table \ref{tab:all_MCalgos_frob}, we compare NORST-smoothing with existing MC solutions (for which code is available). This table displays the Monte-Carlo mean of the normalized Frobenius norm error along with time-taken per column displayed in parentheses. We compare two solvers for nuclear norm min (NNM) -- (i) Singular Value Thresholding (SVT) with maximum iterations as $500$, tolerance as $10^{-8}$, $\delta = 1.2/\rho$, and $\tau = 5 \sqrt{nd}$ and (ii) Inexact Augmented Lagrangian Multiplier (IALM) \cite{ialm_nnm} with maximum iterations $500$ and tolerance $10^{-16}$. We also evaluate the projected Gradient Descent (projected-GD) algorithm of \cite{rmc_gd}, this is a non-convex and hence fast approach, with the best sample complexity among non-convex approaches. This seems to be the only provable non-convex MC approach for which code is available.
NORST-smoothing used $K=33$ and $\alpha=2r$.

The matrix $\L$ was generated as described in Sec. \ref{sec:sims_fixed} for the ``fixed'' subspace rows and as in Sec. \ref{sec:sims_change} (piecewise constant subspace change) for the ``Noisy, Changing'' subspace row. The observed entries set followed the Bernoulli model with different values of $\rho$ in the different rows. The table demonstrates our discussion from Sec. \ref{sec:mainres_noisefree}. (1) In all cases, NORST-smoothing is much faster than both the solvers for convex MC (NNM), but is slower than the best non-convex MC approach (projected-GD). (2) NORST-smoothing is always better than projected-GD (implemented using default code, it is not easy to change the code parameters). It is nearly as good as IALM (one of the two solvers for NNM) when $\rho$ is large, but is worse than IALM when $\rho$ is small.
}

}




\Subsection{Real Video Data} \label{sec:sims_real}
Here we consider the task of Background Recovery for missing data. We use the \texttt{Meeting Room} video which is a benchmark dataset in Background Recovery. It contains $1755$ images of size $64$x$80$ in which a curtain is moving in the wind. Subsequently, there are $1209$ frames in which a person walks into the room, writes on a blackboard, and exits the room. The first $1755$ frames are used for ST-miss while the subsequent frames are used for RST-miss (since we can model the person as a sparse outlier \cite{rpca}).

We generate the set of observed entries using the Bernoulli model with $\rho = 0.9$. In all experiments, we use the estimate of rank as $r=30$. The parameters of NORST-miss are $\alpha = 60$, $K = 3$, and $\lthres = 2 \times 10^{-3}$. We noticed that PETRELS failed to retrieve the background with default parameters so we increased \texttt{max$\_$cycles}$=10$ and refer to this as PETRELS($10$) in the sequel. Furthermore, we also ensured that the input data matrix has more columns than rows by transposing the matrix when necessary. All other algorithms are implemented as done in the previous experiments.
We observed that NORST-miss and SVT provide a good estimate of the background and NORST is $\approx 150$x faster. The relative Frobenius error is provided in the last row of Table. \ref{tab:all_MCalgos_frob}. Notice that, in this case, SVT outperforms IALM and NORST, but NORST is the fastest one.  These results are averaged over $10$ independent trials.

{\bf Moving Object Missing Entries:} In our second video experiment, we generated the set of missing entries using the moving object model with $\rho = 0.98$. All algorithms are implemented as in the previous experiment. Interestingly, even though we observe $98\%$ of the entries, the performance of all algorithms degrade compared to the Bern($0.9$). This is possibly because the support sets are highly correlated over time and thus the assumptions of other algorithms break down. The results are shown in Fig. \ref{fig:vid_mo_st}. Observe that NORST-miss and SVT provide the best visual comparison and NORST-miss is faster than SVT by $\approx 400$x. PETRELS($10$) contains significant artifacts in the recovered background and IALM provides a {\em static} output in which the movements of the curtain are not discernible.

\begin{table}[ht!]
	{
		\caption{\small{Comparing recover error for Robust MC methods. Missing entries were Bernoulli with $\rho = 0.9$, and the outliers were sparse Moving Objects with $\rho_{\sparse} = 0.95$. The time taken per sample is shown in parentheses.}}
		\begin{center}
		\resizebox{.8\linewidth}{!}{
				\begin{tabular}{cc c c }
					\toprule
					{NORST-miss-rob} & GRASTA-RMC & projected-GD\\ \midrule
					${0.0832}$ ($ 3$) & $0.1431$ ($2.9$) & $0.5699$ ($2$)\\
					\bottomrule
				\end{tabular}
				}
		\end{center}
		\label{tab:rmc}
					}
\end{table}

\Subsection{RST-miss and RMC} \label{sec:sims_rmc}
In this experiment, we consider the RST-miss problem, i.e., we generate data according to \eqref{eq:rmc_prob}. We generate the low rank matrix, $\L$, as done in experiment $1$ (single subspace). We generate the sparse matrix, $\S$ as follows: we use the Moving Object Model to generate the support sets such that $s/n = 0.05$ and $b_0 = 0.05$ which translates to $\rho_{\sparse} = 0.05$ {\em fraction of sparse outliers}. The non-zero magnitudes of $\X$ are generated uniformly at random between $[x_{\text{min}}, x_{\text{max}}]$ with $x_{\text{min}} = 10$ and $x_{\text{max}}=25$. We generated the support of observed entries using Bernoulli Model with probability $\rho_{\text{obs}} = 0.9$.

For initialization step of NORST-miss-robust (Algorithm 2), for the first $t_{\train} = 400$ data samples, we set  $(\yt)_i = 10$ for all $i \in \Tmisst$. We do this to allow us to use AltProj \cite{robpca_nonconvex}, which is an RPCA solution, for obtaining the initial subspace estimate.
The parameters for this step are set as $500$ maximum iterations of AltProj, and tolerance $10^{-3}$. The other algorithm parameters for NORST-miss-robust are $\alpha = 60$, $K = 33$, $\lthres = 7.8 \times 10^{-4}$, $\xi = x_{\min}/15$, and $\omega_{supp} = \x_{\text{min}}/2 = 5$. We compare\footnote{we do not compare it with NNM based methods for which code is not available online} GRASTA-RMC \cite{grass_undersampled} and projected-GD \cite{rmc_gd}. For GRASTA-RMC we used the tolerance $10^{-8}$, and \texttt{max$\_$cycles}$=1$. For projected-GD, we use the default tolerance $10^{-1}$ and max. iterations $70$. The results are given in Table. \ref{tab:rmc}. Observe that NORST-miss-robust obtains the best estimate among the RMC algorithms.

{\bf Real video data:} In this experiment, we consider Background recovery applied on the second part of the dataset (last $1209$ frames). In addition to the person who enters the room and writes on the board (sparse component), we generate missing entries from the Bernoulli model with $\rho= 0.9$. We initialize using AltProj with tolerance $10^{-2}$ and $100$ iterations. We set $\omega_{supp,t} = 0.9 \|\yt\|/\sqrt{n}$ using the approach of \cite{rrpcp_icml}. The comparison results are provided in Fig. \ref{fig:vid_rmc}. Notice that both GRASTA-RMC and projected-GD fail to accurately recover the background. Although NORST-miss-robust exhibits certain artifacts around the edges of the sparse object, it is able to capture most of the information in the background.

\section{Conclusions and Open Questions}\label{sec:conc}
This work studied the related problems of subspace tracking in missing data (ST-miss) and its robust version. We show that our proposed approaches are provably accurate under simple assumptions on only the observed data (in case of ST-miss), and on the observed data and initialization (in case of robust ST-miss).
Thus, in both cases, the required assumptions are only on the algorithm inputs, making both results {\em complete guarantees}. Moreover, our guarantees show that our algorithms need near-optimal memory; are as fast as vanilla PCA; and can detect and track subspace changes quickly.
We provided a detailed discussion of related work on (R)ST-miss, (R)MC, and streaming PCA with missing data, that help place our work in the context of what already exists. We also show that  NORST-miss and NORST-miss-robust have good experimental performance as long as the fraction of missing entries is not too large. 

Our guarantee for ST-miss is particularly interesting because it does not require slow subspace change and good initialization. Thus, it can be understood as a novel mini-batch and nearly memory-optimal solution for low-rank Matrix Completion, that works under similar assumptions to standard MC, but needs more numbers of observed entries in general (except in the regime of frequently changing subspaces).

While our approaches have near-optimal memory complexity, they are not streaming. This is because they use SVD and hence need multiple passes over short batches of stored data. A key open question is whether a fully streaming provably correct solution can be developed without assuming the i.i.d. Bernoulli model on the set of missing entries? Two other important open questions include: (i) can the required number of observed entries be reduced (the limiting bound here is the bound on missing fractions per column); and (ii) in case of robust ST-miss, can the lower bound on outlier magnitudes be removed?
{
Another question is whether we can use the tracked estimates for ``control''? For example, can we use the current estimate of the subspace and of the true data vectors to decide how to sample the set of observed entries at the next time instant or later (in applications where one can design this set)?}%


\appendices
\renewcommand{\thetheorem}{\thesection.\arabic{theorem}}

\section{Proof of Theorem \ref{thm:stmiss} and Corollary \ref{cor:noisy}\label{sec:proof_outline}}
This appendix can be shortened/removed after review.
Much of the proof is a simplification of the proof for NORST for RST \cite[Sections 4, 5 and Appendix A]{rrpcp_icml}. The analysis of subspace change detection is exactly the same as done there (see Lemma 4.8 and Appendix A of \cite{rrpcp_icml}) and hence we do not repeat it here. We explain the main ideas of the rest of the proof. To understand it simply, assume that $\that_j=t_j$, i.e, that $t_j$ is known. We use the following simplification of \cite[Remark 2.3]{pca_dd_isit} to analyze the subspace update step. 

\begin{corollary}[PCA in sparse data-dependent noise (Remark 2.3 of \cite{pca_dd_isit})]\label{cor:pca_dd}
For $t = 1, \cdots, \alpha$, suppose that $\yt = \lt + \wt + \vt$ with $\wt= \I_{\Tt}\M_{s,t}\lt$ being sparse noise with support $\T_t$, and  $\lt = \P \at$ where $\P$ is a $n\times r$ basis matrix and $\at$'s satisfy the statistical right-incoherence assumption given in the theorem. Let $\Phat$ be the matrix of top $r$ eigenvectors of $\frac{1}{\alpha} \sum_t \yt \yt{}'$.
Assume that $\max_t \|\M_{s,t} \P\| \leq q$ for a $q \le 3$ and  that the fraction of non-zeros in any row of the matrix $[\w_1, \cdots, \w_{\alpha}]$ is bounded by $\bz$. Pick an $\epsilon_{\mathrm{SE}} >0$. If
$6 \sqrt{\bz} q f + \lambda_v^+ / \lambda^-  < 0.4 \epsilon_{\mathrm{SE}}$ and if $\alpha \ge \alpha^*$ where
\[
\alpha^* := C \max\left( \frac{q^2 f^2}{\epsilon_{\mathrm{SE}}^2} r \log n, \frac{\frac{\lambda_v^+}{\lambda^-} f}{\epsilon_{\mathrm{SE}}^2} r_v \log n\right),
\]
then, w.p. at least $1- 10n^{-10}$, $\SE(\Phat, \P) \le \epsilon_{\mathrm{SE}}$. 
\end{corollary}

First assume that $\vt=0$ so that $\lambda_v^+ = 0$ and $r_v=0$. Also, let  $b_0:= \frac{c_2}{f^2}$ denote the bound on $\missfracrow_\alpha$ assumed in the Theorem.

Using the expression for $\hat\z_t$ given in \eqref{eq:zhatt}, it is easy to see that the error $\et := \lt - \lhatt$ satisfies
\begin{align}\label{eq:etdef}
\et = \I_{\Tmisst} \left( \bpsi_{\Tmisst}{}'\bpsi_{\Tmisst} \right)^{-1} \I_{\Tmisst}{}' \bpsi \lt,
\end{align}
with $\bpsi = \I-\Phat_{(t-1)} \Phat_{(t-1)}{}'$. 
For the first $\alpha$ frames, $\Phat_{(t-1)} = \bm{0}$ (zero initialization) and so, during this time, $\bpsi = \I$.

We need to analyze the subspace update steps one at a time.  We first explain the main ideas of how we do this for $j>0$ and then explain the different approach needed for $j=0$ (because of zero initialization).
Consider a general $j>0$ and $k=1$, i.e., the first subspace update interval of estimating $\P_j$. In this interval  $\bpsi = \I - \Phat_{j-1} \Phat_{j-1}{}'$ and recall that $\Phat_{j-1} = \Phat_{j-1,K}$. Assume that $\SE(\Phat_{j-1}, \P_{j-1}) \le \zz$.

Using the $\mu$-incoherence assumption, the bound on $\missfraccol:= \max_t |\Tmisst|/n$, $\SE(\Phat_{j-1}, \P_{j-1}) \le \zz$ (assumed above),
and recalling from the algorithm that $\Phat_j : = \Phat_{j,K}$, it is not hard to see that\footnote{Use the RIP-denseness lemma from \cite{rrpcp_perf} and some simple linear algebra which includes a triangle inequality type bound for $\SE$. See the proof of item 1 of Lemma 4.7 of \cite{rrpcp_icml}}, for all $j$,
\\ $\SE(\Phat_{j-1}, \P_j) \le \SE(\Phat_{j-1}, \P_{j-1}) + \SE(\P_{j-1}, \P_j)$\\ $\| \I_\Tmisst{}' \P_j\| \le 0.1$,
\\ $\| \I_\Tmisst{}' \Phat_{j,k}\| \le \SE(\Phat_{j,k}, \P_j) +  0.1$,
\\ $\| \I_\Tmisst{}' \Phat_{j-1}\| \le \zz +  0.1$,
\\ $\| \left( \bpsi_{\Tmisst}{}' \bpsi_{\Tmisst}\right)^{-1} \| \le 1.2$ with $\bpsi = \I - \Phat_{j,k} \Phat_{j,k}{}'$.

Next we apply Corollary \ref{cor:pca_dd} to the $\lhatt$'s. This bounds the subspace recovery error for PCA in sparse data-dependent noise.  Since $\lhat_t = \lt + \et$ with $\et$ satisfying \eqref{eq:etdef}, clearly, $\et$ is sparse and dependent on $\lt$ (true data). In the notation of  Corollary \ref{cor:pca_dd}, $\yt \equiv \lhat_t$, $\wt \equiv \et$, $\vt = 0$, $\Tt \equiv \Tmisst$, $\lt \equiv \lt$, $\Phat = \Phat_{j,1}$, $\P = \P_j$, and $\M_{s,t} = -\left( \bpsi_{\Tmisst}{}' \bpsi_{\Tmisst}\right)^{-1} \bpsi_{\Tmisst}{}'$ with $\bpsi = \I - \Phat_{j-1} \Phat_{j-1}{}'$. Thus, using bounds from above, $\norm{\M_{s,t} \P} = \| \left( \bpsi_{\Tmisst}{}' \bpsi_{\Tmisst}\right)^{-1} \I_\Tmisst{}' \bpsi \P_j\| \leq
\| \left( \bpsi_{\Tmisst}{}' \bpsi_{\Tmisst}\right)^{-1} \|  \| \I_\Tmisst{}' \| \| \bpsi \P_j\| \le
1.2 (\zz + \SE(\P_{j-1}, \P_j)) \equiv q$. Also, $\bz \equiv b_0 := \frac{c_2}{f^2}$ ($c_2 = 0.001$) which is the upper bound on $\missfracrow_{\alpha}$ and so $1.2 (\zz + \SE(\P_{j-1}, \P_j)) < 1.2(0.01+\Delta) <  1.3$ since $\Delta \le 1$. Thus $q < 3$.
We apply Corollary \ref{cor:pca_dd} with $\varepsilon_{\mathrm{SE}} = q/4$. All its assumptions hold because we have set $\alpha = C f^2 r \log n$ and because we have let $b_0=0.001/f^2$ and so the required condition $3\sqrt{b} f q \le 0.9 \varepsilon_{\mathrm{SE}} / (1 + \varepsilon_{\mathrm{SE}})$ holds.
We conclude that $\SE(\Phat_{j,1}, \P_j) \le  1.2 (0.01 + \Delta) / 4 = 0.3(0.01+\Delta):= q_1$ whp.

The above is the base case for an induction proof. For the $k$-th subspace update interval, with $k > 1$, we use a similar approach to the one above.  Assume that at the end of the $(k-1)$-th interval, we have $\SE(\Phat_{j,k-1}, \P_j) \le q_{k-1}:= 0.3^{k-1} (0.01 + \Delta)$ whp
In this interval,  $\norm{\M_{s,t} \P} \leq  1.2  \| \I_\Tmisst{}' \|  \| \bpsi \P_j\| \le 1.2 \SE(\Phat_{j,k-1}, \P_j) \le q_{k-1} =  1.2 \cdot 0.3^{k-1} (0.01 + \Delta) \equiv q$.
We apply Corollary \ref{cor:pca_dd} with $\varepsilon_{\mathrm{SE}} = q/4$. This is possible because we have let $b_0=0.001/f^2$ and so the required condition $3\sqrt{b} f q \le 0.9 (q/4) / (1 + q/4)$ holds.  Thus we can  conclude that $\SE(\Phat_{j,k}, \P_j) \le 1.2 \cdot 0.3^{k-1} (0.01 + \Delta)  / 4 = 0.3^{k} (0.01 + \Delta):=q_k$ whp
Thus starting from $\SE(\Phat_{j,k-1}, \P_j) \le q_{k-1}:= 0.3^{k-1} (0.01 + \Delta)$, we have shown that $\SE(\Phat_{j,k}, \P_j) \le   0.3^{k} (0.01 + \Delta)$. This along with the base case, implies that we get $\SE(\Phat_{j,k}, \P_j) \le   0.3^{k} (0.01 + \Delta)$ for all $k=1,2,\dots,K$. The choice of $K$ thus implies that $\SE(\Phat_{j}, \P_j) =\SE(\Phat_{j,K}, \P_j) \le \zz$.

For $j=0$ and first subspace interval ($k=1$), the proof is a little different from that of \cite{rrpcp_icml} summarized above. The reason is we use zero initialization. Thus, in the first update interval for estimating $\P_0$, we have $\bpsi = \I$. In applying the PCA in sparse data-dependent noise result of Corollary \ref{cor:pca_dd}, everything is the same as above except that we now have $\M_{s,t} = \I_\Tmisst{}'$ and so we get $\norm{\M_{s,t} \P} \leq 0.1$. Thus in this case $q =0.1 < 3$. The rest of the argument is the same as above.

Now consider $\vt \neq 0$. Recall that the effective noise dimension of $\vt$ is $r_v = \max_t \|\vt\|^2/\lambda_v^+$ where $\lambda_v^+ = \|\E[\vt\vt{}']\|$. Furthermore, recall that $\epsilon_{\mathrm{SE}} = q/4$. Thus, in order to obtain $\zz$-accurate estimate in the noisy case, we will require that $\alpha = \mathcal{O}\left( \max\left( f^2 r \log n, \frac{\frac{\lambda_v^+}{\lambda^-} f r_v \log n}{\epsilon_{\mathrm{SE}}^2} \right)\right)$. Thus, we set $\epsilon_{\mathrm{SE}} = c \sqrt{{\lambda_v^+}/{\lambda^-}}$ to ensure that the dependence on $\zz$ is on logarithmic (that comes from expression for $K$).

The above provides the basic proof idea in a condensed fashion but does not define events that one conditions on for each interval, and also does not specify the probabilities. For all these details, please refer to Sections IV and V and Appendix A of \cite{rrpcp_icml}.

\section{Proof of Corollary \ref{cor:dyn_rmc}}\label{sec:proof_rmc}
This proof is also similar to that of NORST for RST \cite{rrpcp_icml}. The difference is  NORST-miss-robust uses noisy modified CS \cite{modcsjournal,stab_jinchun_jp} to replace $l_1$ min. In comparison to the ST-miss proof summarized above,  we also have to deal with arbitrary outliers, in addition to missing data. 
This uses requires sparse support recovery with partial subspace knowledge. This is solved by modified-CS followed by thresholding based support recovery. To bound the modified-CS error, we apply Lemma 2.7 of \cite{stab_jinchun_jp}.  This uses a bound on $\|\bt\| = \|\bpsi \lt\|$ and a bound on  the $(\missfraccol \cdot n + 2 \outfraccol \cdot n)$-RIC of $\bpsi$. We obtain both these exactly as done for \cite[Lemma 4.7, Item 1]{rrpcp_icml}: the former uses the slow subspace change bound and the boundedness of $\at$; for the latter we use the $\mu$-incoherence/denseness assumption and bounds on $\outfraccol$ and $\missfraccol$, and the RIP-denseness lemma of \cite{rrpcp_perf}.
With the modified-CS error bound, we prove exact support recovery using the lower bound on $\xmint$. algorithm parameter values of $\xi$ and $\omega_{supp}$.


{

}

%

\bibliographystyle{IEEEbib}
\bibliography{../bib/tipnewpfmt_kfcsfullpap}

\begin{thebibliography}{10}

\bibitem{rst_miss_icassp}
P.~Narayanamurthy, V.~Daneshpajooh, and N.~Vaswani,
\newblock ``Provable memory-efficient online robust matrix completion,''
\newblock in {\em IEEE Int. Conf. Acoust., Speech and Sig. Proc. (ICASSP)}.
  IEEE, 2019, pp. 7918--7922.

\bibitem{st_miss_isit}
P.~Narayanamurthy, V.~Daneshpajooh, and N.~Vaswani,
\newblock ``Provable subspace tracking with missing entries,''
\newblock {\em to appear, ISIT}, 2019.

\bibitem{rrpcp_icml}
P.~Narayanamurthy and N.~Vaswani,
\newblock ``Nearly optimal robust subspace tracking,''
\newblock in {\em International Conference on Machine Learning}, 2018, pp.
  3701--3709.

\bibitem{golubtracking}
P.~Comon and G.~H. Golub,
\newblock ``Tracking a few extreme singular values and vectors in signal
  processing,''
\newblock {\em Proceedings of the IEEE}, vol. 78, no. 8, pp. 1327--1343, 1990.

\bibitem{chi_review}
L.~Balzano, Y.~Chi, and Y.~M. Lu,
\newblock ``Streaming pca and subspace tracking: The missing data case,''
\newblock {\em Proceedings of IEEE}, 2018.

\bibitem{sslearn_jmlr}
A.~Gonen, D.~Rosenbaum, Y.~C. Eldar, and S.~Shalev-Shwartz,
\newblock ``Subspace learning with partial information,''
\newblock {\em The Journal of Machine Learning Research}, vol. 17, no. 1, pp.
  1821--1841, 2016.

\bibitem{rrpcp_proc}
N.~Vaswani and P.~Narayanamurthy,
\newblock ``Static and dynamic robust pca and matrix completion: A review,''
\newblock {\em Proceedings of the IEEE}, vol. 106, no. 8, pp. 1359--1379, 2018.

\bibitem{past}
B.~Yang,
\newblock ``Projection approximation subspace tracking,''
\newblock {\em IEEE Trans. Sig. Proc.}, pp. 95--107, 1995.

\bibitem{past_conv}
B.~Yang,
\newblock ``Asymptotic convergence analysis of the projection approximation
  subspace tracking algorithms,''
\newblock {\em Signal Processing}, vol. 50, pp. 123--136, 1996.

\bibitem{petrels}
Y.~Chi, Y.~C. Eldar, and R.~Calderbank,
\newblock ``Petrels: Parallel subspace estimation and tracking by recursive
  least squares from partial observations,''
\newblock {\em IEEE Trans. Sig. Proc.}, December 2013.

\bibitem{grouse}
L.~Balzano, B.~Recht, and R.~Nowak,
\newblock ``{Online Identification and Tracking of Subspaces from Highly
  Incomplete Information},''
\newblock in {\em Allerton Conf. Comm., Control, Comput.}, 2010.

\bibitem{local_conv_grouse}
L.~Balzano and S.~Wright,
\newblock ``Local convergence of an algorithm for subspace identification from
  partial data,''
\newblock {\em Found. Comput. Math.}, vol. 15, no. 5, 2015.

\bibitem{grouse_global}
D.~Zhang and L.~Balzano,
\newblock ``Global convergence of a grassmannian gradient descent algorithm for
  subspace estimation,''
\newblock in {\em AISTATS}, 2016.

\bibitem{grouse_enh}
G.~Ongie, D.~Hong, D.~Zhang, and L.~Balzano,
\newblock ``Enhanced online subspace estimation via adaptive sensing,''
\newblock in {\em Asilomar}, 2018.

\bibitem{petrels_new}
C.~Wang, Y.~C. Eldar, and Y.~M. Lu,
\newblock ``Subspace estimation from incomplete observations: A
  high-dimensional analysis,''
\newblock {\em JSTSP}, 2018.

\bibitem{streamingpca_miss}
I.~Mitliagkas, C.~Caramanis, and P.~Jain,
\newblock ``Streaming pca with many missing entries,''
\newblock {\em Preprint}, 2014.

\bibitem{eldar_jmlr_ss}
A.~Gonen, D.~Rosenbaum, Y.~C. Eldar, and S.~Shalev-Shwartz,
\newblock ``Subspace learning with partial information,''
\newblock {\em Journal of Machine Learning Research}, vol. 17, no. 52, pp.
  1--21, 2016.

\bibitem{rrpcp_allerton}
C.~Qiu and N.~Vaswani,
\newblock ``Real-time robust principal components' pursuit,''
\newblock in {\em Allerton Conf. on Communication, Control, and Computing},
  2010.

\bibitem{rrpcp_perf}
C.~Qiu, N.~Vaswani, B.~Lois, and L.~Hogben,
\newblock ``Recursive robust pca or recursive sparse recovery in large but
  structured noise,''
\newblock {\em IEEE Trans. Info. Th.}, pp. 5007--5039, August 2014.

\bibitem{rrpcp_aistats}
J.~Zhan, B.~Lois, H.~Guo, and N.~Vaswani,
\newblock ``{Online (and Offline) Robust PCA: Novel Algorithms and Performance
  Guarantees},''
\newblock in {\em Intnl. Conf. Artif. Intell. Stat. (AISTATS)}, 2016.

\bibitem{rrpcp_dynrpca}
P.~Narayanamurthy and N.~Vaswani,
\newblock ``Provable dynamic robust pca or robust subspace tracking,''
\newblock {\em IEEE Transactions on Information Theory}, vol. 65, no. 3, pp.
  1547--1577, 2019.

\bibitem{matcomp_candes}
E.~J. Candes and B.~Recht,
\newblock ``Exact matrix completion via convex optimization,''
\newblock {\em Found. of Comput. Math}, , no. 9, pp. 717--772, 2008.

\bibitem{lowrank_altmin}
P.~Netrapalli, P.~Jain, and S.~Sanghavi,
\newblock ``Low-rank matrix completion using alternating minimization,''
\newblock in {\em STOC}, 2013.

\bibitem{rmc_gd}
Y.~Cherapanamjeri, K.~Gupta, and P.~Jain,
\newblock ``Nearly-optimal robust matrix completion,''
\newblock {\em ICML}, 2016.

\bibitem{matcomp_first}
M.~Fazel,
\newblock ``Matrix rank minimization with applications,''
\newblock {\em PhD thesis, Stanford Univ}, 2002.

\bibitem{recht_mc_simple}
B.~Recht,
\newblock ``A simpler approach to matrix completion,''
\newblock {\em Journal of Machine Learning Research}, vol. 12, no. Dec, pp.
  3413--3430, 2011.

\bibitem{robpca_nonconvex}
P.~Netrapalli, U~N Niranjan, S.~Sanghavi, A.~Anandkumar, and P.~Jain,
\newblock ``Non-convex robust pca,''
\newblock in {\em NIPS}, 2014.

\bibitem{pca_dd_isit}
N.~Vaswani and P.~Narayanamurthy,
\newblock ``Pca in sparse data-dependent noise,''
\newblock in {\em ISIT}, 2018, pp. 641--645.

\bibitem{candes_rip}
E.~Candes,
\newblock ``The restricted isometry property and its implications for
  compressed sensing,''
\newblock {\em C. R. Math. Acad. Sci. Paris Serie I}, 2008.

\bibitem{rpca}
E.~J. Cand{\`e}s, X.~Li, Y.~Ma, and J.~Wright,
\newblock ``Robust principal component analysis?,''
\newblock {\em J. ACM}, vol. 58, no. 3, 2011.

\bibitem{leeb_non_iso}
W.~Leeb and E.~Romanov,
\newblock ``Optimal spectral shrinkage and pca with heteroscedastic noise,''
\newblock {\em arXiv:1811.02201}, €œ2018.

\bibitem{coherent_mc}
Y.~Chen, S.~Bhojanapalli, S.~Sanghavi, and R.~Ward,
\newblock ``Coherent matrix completion,''
\newblock in {\em International Conference on Machine Learning}, 2014, pp.
  674--682.

\bibitem{optspace}
R.H. Keshavan, A.~Montanari, and S.~Oh,
\newblock ``Matrix completion from a few entries,''
\newblock {\em IEEE Trans. Info. Th.}, vol. 56, no. 6, pp. 2980--2998, 2010.

\bibitem{onlineMC1}
C.~Jin, S.~M. Kakade, and P.~Netrapalli,
\newblock ``Provable efficient online matrix completion via non-convex
  stochastic gradient descent,''
\newblock in {\em NIPS}, 2016, pp. 4520--4528.

\bibitem{laura_subspace_match}
L.~Balzano, B.~Recht, and R.~Nowak,
\newblock ``High-dimensional matched subspace detection when data are
  missing,''
\newblock in {\em ISIT}, 2010, pp. 1638--1642.

\bibitem{ojasimplified}
E.~Oja,
\newblock ``Simplified neuron model as a principal component analyzer,''
\newblock {\em Journal of mathematical biology}, vol. 15, no. 3, pp. 267--273,
  1982.

\bibitem{onlineMC2}
A.~Krishnamurthy and A.~Singh,
\newblock ``Low-rank matrix and tensor completion via adaptive sampling,''
\newblock in {\em NIPS}, 2013, pp. 836--844.

\bibitem{mc_luo}
R.~Sun and Z-Q. Luo,
\newblock ``Guaranteed matrix completion via non-convex factorization,''
\newblock {\em IEEE Trans. Info. Th.}, vol. 62, no. 11, pp. 6535--6579, 2016.

\bibitem{lowrank_altmin_no_kappa}
M.~Hardt and M.~Wootters,
\newblock ``Fast matrix completion without the condition number,''
\newblock in {\em COLT}, 2014.

\bibitem{fastmc}
P.~Jain and P.~Netrapalli,
\newblock ``Fast exact matrix completion with finite samples,''
\newblock in {\em Conference on Learning Theory}, 2015, pp. 1007--1034.

\bibitem{ge_1}
R.~Ge, J.~D. Lee, and T.~Ma,
\newblock ``Matrix completion has no spurious local minimum,''
\newblock in {\em NIPS}, 2016, pp. 2973--2981.

\bibitem{ge_best}
R.~Ge, C.~Jin, and Y.~Zheng,
\newblock ``No spurious local minima in nonconvex low rank problems: A unified
  geometric analysis,''
\newblock {\em arXiv preprint arXiv:1704.00708}, 2017.

\bibitem{rmc_altproj}
X.~Jiang, Z.~Zhong, X.~Liu, and H.~C. So,
\newblock ``Robust matrix completion via alternating projection,''
\newblock {\em IEEE Signal Processing Letters}, vol. 24, no. 5, pp. 579--583,
  2017.

\bibitem{mc_altproj}
M.J. Lai and A.~Varghese,
\newblock ``On convergence of the alternating projection method for matrix
  completion and sparse recovery problems,''
\newblock {\em arXiv preprint arXiv:1711.02151}, 2017.

\bibitem{noniid_mc}
S.~Foucart, D.~Needell, Y.~Plan, and M.~Wootters,
\newblock ``De-biasing low-rank projection for matrix completion,''
\newblock in {\em Wavelets and Sparsity XVII}. International Society for Optics
  and Photonics, 2017, vol. 10394, p. 1039417.

\bibitem{universal_mc}
S.~Bhojanapalli and P.~Jain,
\newblock ``Universal matrix completion,''
\newblock in {\em International Conference on Machine Learning}, 2014, pp.
  1881--1889.

\bibitem{modcsjournal}
N.~Vaswani and W.~Lu,
\newblock ``Modified-{CS}: Modifying compressive sensing for problems with
  partially known support,''
\newblock {\em IEEE Trans. Signal Processing}, September 2010.

\bibitem{stab_jinchun_jp}
J.~Zhan and N.~Vaswani,
\newblock ``Time invariant error bounds for modified-{CS} based sparse signal
  sequence recovery,''
\newblock {\em IEEE Trans. Info. Th.}, vol. 61, no. 3, pp. 1389--1409, 2015.

\bibitem{grass_undersampled}
J.~He, L.~Balzano, and A.~Szlam,
\newblock ``Incremental gradient on the grassmannian for online foreground and
  background separation in subsampled video,''
\newblock in {\em IEEE Conf. on Comp. Vis. Pat. Rec. (CVPR)}, 2012.

\bibitem{chouvardas2015robust}
S.~Chouvardas, Y.~Kopsinis, and S.~Theodoridis,
\newblock ``Robust subspace tracking with missing entries: a set--theoretic
  approach,''
\newblock {\em IEEE Trans. Sig. Proc.}, vol. 63, no. 19, pp. 5060--5070, 2015.

\bibitem{mansour_robust_ss_track}
H.~Mansour and X.~Jiang,
\newblock ``A robust online subspace estimation and tracking algorithm,''
\newblock in {\em ICASSP}, 2015, pp. 4065--4069.

\bibitem{ranksparSanghavi}
Y.~Chen, A.~Jalali, S.~Sanghavi, and C.~Caramanis,
\newblock ``Low-rank matrix recovery from errors and erasures,''
\newblock {\em IEEE Trans. Inform. Theory}, vol. 59(7), pp. 4324--4337, 2013.

\bibitem{rpca_gd}
X.~Yi, D.~Park, Y.~Chen, and C.~Caramanis,
\newblock ``Fast algorithms for robust pca via gradient descent,''
\newblock in {\em NIPS}, 2016.

\bibitem{cgls}
C.~C. Paige and M.~A. Saunders,
\newblock ``Lsqr: An algorithm for sparse linear equations and sparse least
  squares,''
\newblock {\em ACM Transactions on Mathematical Software (TOMS)}, vol. 8, no.
  1, pp. 43--71, 1982.

\bibitem{ialm_nnm}
Z.~Lin, M.~Chen, and Y.~Ma,
\newblock ``The augmented lagrange multiplier method for exact recovery of
  corrupted low-rank matrices,''
\newblock {\em arXiv preprint arXiv:1009.5055}, 2010.

\end{thebibliography}



\end{document}